\documentclass{article} 
\usepackage{iclr2027_conference,times}

\usepackage{amsmath,amsfonts,bm}

\def\eqref#1{equation~\ref{#1}}

\def\1{\bm{1}}

\DeclareMathAlphabet{\mathsfit}{\encodingdefault}{\sfdefault}{m}{sl}
\SetMathAlphabet{\mathsfit}{bold}{\encodingdefault}{\sfdefault}{bx}{n}

\usepackage{hyperref}
\usepackage{url}

\usepackage{xcolor}
\usepackage{amsthm}
\newtheorem{theorem}{Theorem}
\newtheorem{corollary}[theorem]{Corollary}
\usepackage{algorithm}
\usepackage{algorithmic}
\usepackage{enumitem}
\usepackage{booktabs}
\usepackage{graphicx}
\usepackage{subfigure}
\usepackage{subcaption}
\usepackage[table]{xcolor}
\definecolor{bestblue}{RGB}{220,235,247}
\title{Beyond Policy Support: Interaction Constrained Offline Reinforcement Learning for Autonomous Driving}

\author{%
Mahmoud Selim\textsuperscript{1,2} \qquad
\textbf{Cristina Cipriani}\textsuperscript{1} \qquad
\textbf{Karl H.~Johansson}\textsuperscript{2} \\
\textsuperscript{1}TRATON, \texttt{\{mahmoud.selim, cristina.cipriani\}@se.traton.com} \\
\textsuperscript{2}KTH Royal Institute of Technology, \texttt{\{mase2, kallej\}@kth.se}
}

\newcommand{\update}[1]{\textcolor{black}{#1}}
\iclrfinalcopy 
\begin{document}

\maketitle

\begin{abstract}

Offline reinforcement learning enables reward-driven policy improvement from fixed datasets without requiring online exploration, making it particularly attractive in safety-critical domains. A central challenge, however, is distribution shift: policy optimization may favor actions that are weakly supported by the offline data, rendering value estimates unreliable. Existing approaches primarily control this shift in the policy's own action space. In interactive environments such as autonomous driving, this can be insufficient: a candidate ego trajectory may remain well supported under the marginal behavior distribution while being poorly supported jointly with the surrounding-agent behavior observed in the logged interaction. We refer to this degradation in interaction support as \emph{interaction distribution shift} (IDS), and introduce \emph{Interaction-Constrained Drive Policy} (ICDP), an offline reinforcement learning framework that explicitly controls interaction-level distribution shift. Starting from the joint data distribution over ego and surrounding-agent futures, we show that joint-support degradation decomposes exactly into an ego-support component and a residual interaction-support component. We recover the latter through contrastive density-ratio estimation, isolating interaction compatibility without explicit joint-density modeling, surrounding-agent prediction, or rollouts in reactive simulators or learned world models during policy optimization. 
Closed-loop evaluations on nuPlan, InterPlan and real-world truck experiments show that ICDP suppresses high-value yet interaction-unsupported trajectory selections and improves performance in interaction-critical driving scenarios.
Project webpage: \url{https://mahmoud-selim.github.io/ICDP/}

\end{abstract}

\section{Introduction} \label{sec:intro}

Autonomous driving requires motion planners that can produce safe, comfortable, and efficient behavior while continuously interacting with other road users. Classical rule- and optimization-based systems provide explicit structure and constraint handling, but often rely on carefully engineered objectives, interaction models, and scenario-specific design, making it difficult to scale their behavior across the long tail of complex traffic situations~\citep{fan2018baidu,hawke2020urban,dauner2023parting}. Learning-based planning offers a compelling alternative by shifting this complexity from manual design to data. Imitation learning (IL), in particular, has enabled driving policies to acquire realistic behavior from large-scale demonstrations~\citep{bojarski2016end,chitta2022transfuser,cheng2024rethinking}, while recent diffusion- and flow-based planners further capture rich multimodal trajectory distributions and generate diverse, high-quality motion plans~\citep{zheng2025diffusion,huang2023gameformer,cheng2024pluto}. Yet greater generative expressiveness does not remove a fundamental limitation of imitation: the policy is still optimized to reproduce the behavior contained in the data, rather than to improve it according to long-horizon driving objectives. Consequently, biases and suboptimal decisions in the demonstrations can be inherited by the learned planner, motivating reward-driven policy optimization beyond imitation~\citep{bansal2018chauffeurnet,cheng2024rethinking}.

Reinforcement learning (RL) provides a natural mechanism beyond imitation by optimizing policies directly for long-horizon driving objectives. Applying RL to autonomous driving, however, is difficult: online methods require extensive closed-loop interaction with sufficiently realistic environments~\citep{kendall2019learning,yang2026raw2drive,zhang2025carplanner}, while model-based approaches shift this burden to learned dynamics or world models, whose inaccuracies can propagate into policy optimization~\citep{wang2025adawm,yang2026raw2drive}. Offline RL offers an appealing alternative by learning exclusively from previously collected driving data, avoiding repeated environment interaction while using bootstrapped value learning to propagate reward beyond the demonstrated behavior~\citep{levine2020offline,noguchi2026pseudo}. This benefit comes with a fundamental challenge: \emph{distribution shift}. As policy optimization moves toward actions that are weakly represented in the dataset, the critic must extrapolate beyond its training support and can assign unreliable values to out-of-distribution behavior. Existing offline RL methods therefore control policy improvement through conservative value estimation, in-sample learning, or explicit regularization toward the behavior distribution~\citep{kumar2020conservative,kostrikov2021offline,peng2019advantage,wang2022diffusion,fang2025diffusion,huang2025lane}. Yet these mechanisms largely characterize support in the policy's own action space. In interactive driving, this is not enough: an ego trajectory can itself be well supported by the data while being poorly supported \emph{jointly} with the surrounding-agent evolution observed in the same interaction. We refer to this loss of interaction support as \emph{interaction distribution shift} (IDS). This raises the central question of our work: \emph{How can an offline driving policy improve beyond the demonstrations while remaining supported not only as an ego trajectory, but as part of the multi-agent interaction represented in the data?}

We answer this question with \textsc{ICDP} (\emph{Interaction-Constrained Drive Policy}), an offline RL framework for reward-driven improvement preserving support at the level of multi-agent interactions. \textsc{ICDP} combines a trajectory-level diffusion policy with chunk-level value learning to evaluate the long-horizon utility of candidate ego trajectories. Its key idea is to go beyond conventional ego-trajectory regularization and explicitly control how policy updates alter the support of the logged ego--agent interaction. We show that the degradation of joint interaction support decomposes exactly into an ego-support term and a residual interaction-support term corresponding to interaction distribution shift. Crucially, the latter can be recovered through contrastive density-ratio estimation, allowing \textsc{ICDP} to assess whether a candidate ego trajectory remains compatible with the surrounding-agent behavior represented in the data without explicitly modeling the high-dimensional joint trajectory density, predicting agent responses, or performing reactive simulator or world-model rollouts during policy optimization. Our contributions are: 
\begin{itemize}[
    leftmargin=1.3em,
    itemsep=4pt,
    topsep=2pt,
    parsep=0pt,
    partopsep=0pt
]
\item We identify and formalize \emph{interaction distribution shift} (IDS), a source of distribution shift specific to interactive offline RL: an ego trajectory may remain well supported marginally while losing support jointly with the surrounding-agent evolution. 

\item We derive a contrastive density-ratio estimator whose residual score recovers the interaction-support component of IDS without explicit joint-density estimation or prediction of surrounding-agent futures, and integrate this constraint with chunk-level value learning and diffusion-policy optimization in \textsc{ICDP}.

\item We evaluate \textsc{ICDP} in closed-loop driving scenarios on nuPlan and InterPlan, as well as in real-world truck experiments, demonstrating its ability to suppress high-value but interaction-unsupported trajectory selections and improve policy performance in challenging scenarios.
\end{itemize}
\section{Background \& Related Works} \label{sec:related}

\paragraph{Learning-based and generative motion planning.}
Imitation learning (IL) has become a dominant paradigm for autonomous-driving planning, progressing from rasterized representations and convolutional architectures~\cite{bojarski2016end,codevilla2019exploring} to structured vector representations and transformers that better capture scene context and interactions~\cite{liang2020learning,casas2021mp3,chitta2022transfuser,hu2023planning,jiang2023vad,cheng2024rethinking}. More recently, diffusion-based generative planners have enabled multimodal trajectory generation and flexible inference-time guidance~\cite{zheng2025diffusion,liao2025diffusiondrive}. While these advances increase policy expressiveness, improvement remains largely driven by demonstrated behavior. In contrast, we use a generative trajectory policy as the foundation for reward-driven improvement from offline interaction data.

\paragraph{Reinforcement learning for autonomous driving.}
RL complements imitation-based planning by optimizing driving policies directly for long-horizon reward. Model-free approaches learn driving behavior through closed-loop environment interaction~\cite{toromanoff2020end}, while model-based RL improves training efficiency by learning environment dynamics and optimizing policies through imagined rollouts~\cite{li2024think2drive,wang2025adawm}. Recent work has further demonstrated the scalability of RL-based planning: CarPlanner combines autoregressive trajectory generation with expert-guided rewards for large-scale policy optimization~\cite{zhang2025carplanner}, while CaRL scales PPO with simple reward design across CARLA and nuPlan~\cite{jaeger2025carl}. Interaction-aware RL has also incorporated explicit intent inference and multi-agent policy learning to reason about surrounding drivers~\cite{wu2023iplan}. Collectively, these methods demonstrate the potential of reward-driven planning, but typically obtain counterfactual feedback through repeated environment interaction or learned dynamics. In contrast, our setting considers policy improvement entirely from fixed driving logs, without simulator interaction or model-generated traffic evolution during optimization.

\paragraph{Offline RL and support constraints.}
Offline RL learns reward-maximizing policies from fixed datasets, but distribution shift can cause value overestimation on poorly supported actions. Existing methods mitigate this through conservative value estimation, behavior or support constraints, and in-sample policy extraction~\cite{kumar2020conservative,kumar2019stabilizing,fujimoto2021minimalist,kostrikov2021offline}. These ideas have been extended to diffusion policies~\cite{wang2022diffusion,fang2025diffusion,gao2025behavior,selimnoisy} and temporally extended actions~\cite{li2026reinforcement}. In autonomous driving, offline RL has been studied through model-based, sequence-modeling, traffic-generation, and behavior-regularized approaches~\cite{diehl2021umbrella,li2023uncertainty,rowe2024ctrl,noguchi2026pseudo}, including explicit support constraints for lane changing~\cite{huang2025lane}. Offline multi-agent methods similarly address joint-action distribution shift~\cite{nguyen2025comadice}, but optimize multiple learning agents. In contrast, \textsc{ICDP} optimizes only the ego policy, treating surrounding traffic as logged environment evolution, and decomposes joint interaction-support degradation into complementary ego-trajectory and conditional interaction-support components.

\section{Problem Formulation}
We formulate autonomous driving planning as a discounted Markov decision process (MDP) $\mathcal{M} = (\mathcal{S}, \mathcal{A}, r, \rho, p, \gamma)$, where $s_t\in\mathcal{S}$ denotes the current driving scene, $a_t\in\mathcal{A}$ is the primitive ego control, $r(s,a)$ is the reward function, $\rho(s)$ the initial state distribution, $p(s'|s,a)$ the transition dynamics, and $\gamma \in [0,1)$ the discount factor. Rather than learning over isolated controls, we operate in a temporally extended action space and define an ego-action chunk
\begin{equation}
    \tau_{e,t}
    =
    (a_t,\ldots,a_{t+H-1})
    \in \mathcal{A}^{H},
\end{equation}
over a committed horizon $H$. Accordingly, the policy and critic are defined as
$\pi_{\theta}(\tau_{e,t}\mid s_t),$ and $Q_{\phi}(s_t,\tau_{e,t}),$
where the critic estimates the return obtained by executing the complete chunk. The offline dataset is represented as
\begin{equation}
    \mathcal{D}
    =
    \left\{
        s_t,\,
        \tau_{e,t}^{\beta},\,
        \tau_{a,t}^{\beta},\,
        R_t^{(H)},\,
        s_{t+H}
    \right\},
\end{equation}
where $\tau_{e,t}^{\beta}$ and $\tau_{a,t}^{\beta}$ are the logged ego-action chunk and the corresponding surrounding-agent futures, and $ R_t^{(H)}=\sum_{k=0}^{H-1}\gamma^k r_{t+k}.$
The critic is trained with the chunk-level target:
\begin{equation}
    y_t
    =
    R_t^{(H)}
    +
    \gamma^H
    Q_{\bar{\phi}}
    \left(
        s_{t+H},
        \hat{\tau}_{e,t+H}
    \right),
    \quad
    \hat{\tau}_{e,t+H}
    \sim
    \pi_{\theta}(\cdot\mid s_{t+H}).
\end{equation}
where $Q_{\bar{\phi}}$ is a delayed target critic. Because $Q_{\phi}$ conditions on the complete action sequence that generated the intermediate rewards, this backup avoids the off-policy bias incurred when an $H$-step return is applied to a single-action critic~\cite{li2026reinforcement}. The surrounding-agent trajectories remain part of the observed environment evolution and are later used to characterize interaction support.

A standard behavior-regularized offline RL objective
improves the policy with respect to the learned critic while restricting
its deviation from the data distribution:
\begin{align}
    \max_{\theta}\quad
    &\mathbb{E}_{\substack{s\sim d_{\mathcal D}\\
                           \hat{\tau}_e\sim\pi_{\theta}(\cdot\mid s)}}
    \left[
        Q_{\phi}(s,\hat{\tau}_e)
    \right]
    \nonumber\\
    \text{s.t.}\quad
    &\mathbb{E}_{s\sim d_{\mathcal D}}
    \left[
        D\!\left(
            \pi_{\theta}(\cdot\mid s),
            \beta(\cdot\mid s)
        \right)
    \right]
    \leq \epsilon_e ,
\end{align}
where $d_{\mathcal D}$ is the empirical state distribution and $D$
denotes a suitable divergence or behavior-regularization measure. While this constraint limits distribution shift in the ego-action
space, it does not ensure that a candidate ego trajectory remains
compatible with the surrounding-agent evolution represented in the
same driving interaction. In particular, high ego support,
$p_{\mathcal D}(\hat{\tau}_e\mid s)$, does not necessarily imply high
joint support,
$p_{\mathcal D}(\hat{\tau}_e,\tau_a^\beta\mid s)$.
Given the interactive nature of driving, our objective is therefore to maximize long-horizon return while
controlling distribution shift in the joint
ego--agent interaction space $p_{\mathcal D}(\hat{\tau}_e,\tau_a^\beta\mid s)$.

\section{Methodology}

In this section, we introduce \textsc{ICDP}, an offline RL framework for diffusion-based motion planning that improves long-horizon driving performance while preserving support at the level of multi-agent interactions. \textsc{ICDP} combines chunk-level value learning with interaction-aware policy regularization, explicitly controlling both deviation from the logged ego-trajectory distribution and \emph{interaction distribution shift} (IDS) induced by policy improvement. We first present the overall actor--critic framework, then derive a tractable characterization of IDS together with a contrastive estimator, and finally incorporate the resulting interaction constraint into diffusion-policy optimization.


\subsection{\textsc{ICDP} Overview}
\label{sec:icdp_overview}

\begin{figure*}[t]
    \centering
    \includegraphics[width=\textwidth,height=5.4cm]{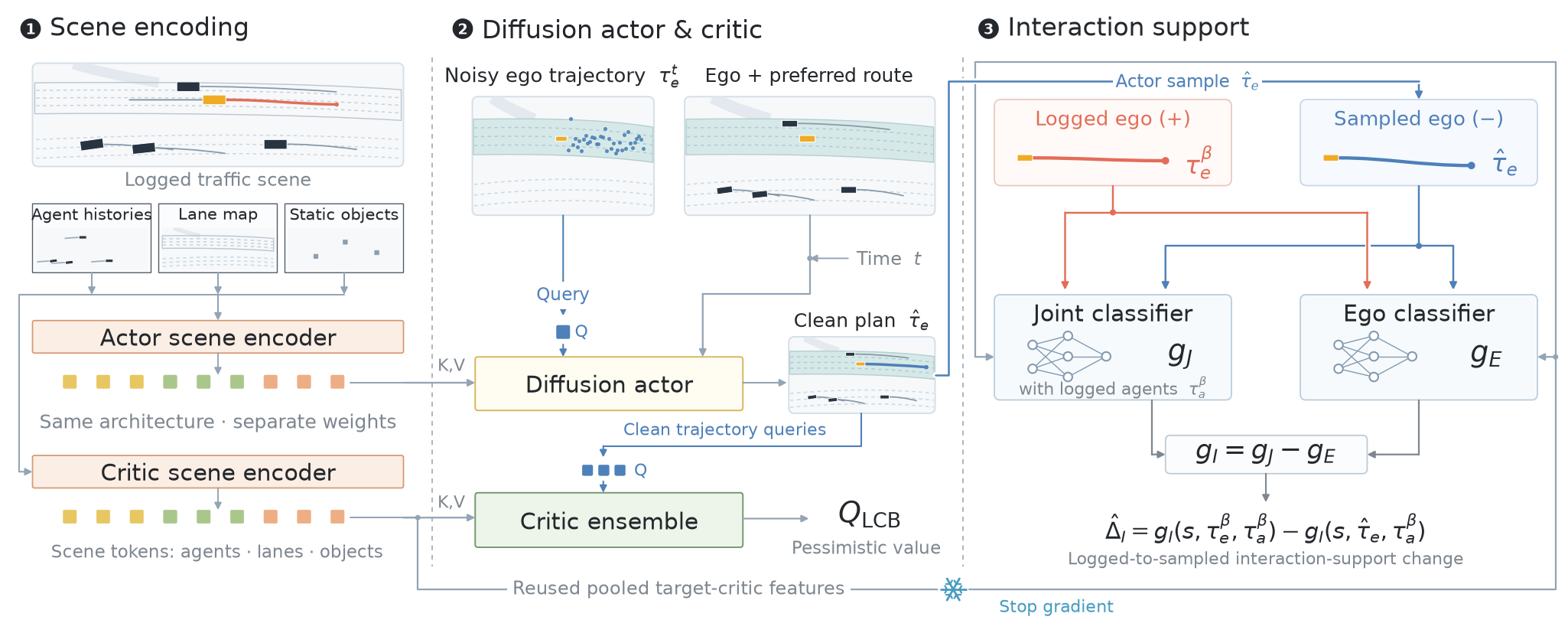}
    \caption{
    Overview of \textsc{ICDP}. The logged scene is encoded into structured
    scene features used by the diffusion actor and critic. The actor generates
    candidate ego trajectories, which are evaluated by a pessimistic
    chunk-level critic and by the interaction-support module. The latter
    combines a joint ego--agent classifier with an ego-only classifier; their
    residual score isolates interaction-specific support and yields the IDS
    estimate used to constrain policy improvement.
    }\vspace{-3mm}
    \label{fig:icdp_architecture}
\end{figure*}

\textsc{ICDP} improves a diffusion-based driving policy from fixed interaction
logs while constraining policy improvement to remain supported by the
multi-agent interactions represented in the dataset. As illustrated in
Fig.~\ref{fig:icdp_architecture}, the framework consists of three components:
a scene representation shared in structure across the actor and critic, a
diffusion actor with an ensemble of chunk-level critics for value-guided
policy improvement, and an interaction-support module that evaluates the
compatibility of policy-generated ego trajectories with the logged
surrounding-agent evolution. The actor proposes candidate ego trajectories,
while the critic and interaction-support estimator operate on the
corresponding clean trajectories to provide value and support
signals for policy optimization.

Given a logged interaction
$(s,\tau_e^\beta,\tau_a^\beta)\sim\mathcal{D}$, where $\tau_e^\beta$ and
$\tau_a^\beta$ denote the logged ego and surrounding-agent trajectories,
respectively, the policy generates a candidate ego-action chunk
\begin{equation}
    \hat{\tau}_e \sim \pi_\theta(\cdot\mid s).
    \label{eq:candidate_trajectory}
\end{equation}
The candidate is evaluated by an ensemble of critics
$\{Q_{\phi_k}(s,\tau_e)\}_{k=1}^{K}$. To reduce exploitation of uncertain
value estimates during policy improvement, we use the lower-confidence-bound
estimate
\begin{equation}
    Q_{\mathrm{LCB}}(s,\tau_e)
    =
    \frac{1}{K}\sum_{k=1}^{K} Q_{\phi_k}(s,\tau_e)
    -
    \kappa_Q
    \operatorname{Std}_{k}
    \left[
        Q_{\phi_k}(s,\tau_e)
    \right],
    \label{eq:q_lcb}
\end{equation}
where $\kappa_Q\geq 0$ controls the degree of pessimism.
Maximizing value alone, however, does not guarantee reliable offline policy
improvement. A candidate ego trajectory may receive a high value estimate while
inducing an ego--agent interaction that is weakly supported by the logged data.
We therefore seek policy improvements that increase long-horizon return while
preserving support under the joint distribution of ego and surrounding-agent
trajectories.

To formalize this requirement, consider replacing the logged ego trajectory
$\tau_e^\beta$ with a policy-generated candidate $\hat{\tau}_e$ while retaining
the surrounding-agent trajectory $\tau_a^\beta$ observed in the same
interaction. We define the resulting \emph{joint-support degradation} as
\begin{equation}
    \Delta_J
    \left(
        s,\tau_e^\beta,\hat{\tau}_e,\tau_a^\beta
    \right)
    =
    \log
    \frac{
        p_{\mathcal D}
        (\tau_e^\beta,\tau_a^\beta\mid s)
    }{
        p_{\mathcal D}
        (\hat{\tau}_e,\tau_a^\beta\mid s)
    }.
    \label{eq:joint_support_degradation}
\end{equation}
A positive value of $\Delta_J$ indicates that the interaction formed by the
candidate ego trajectory is less supported by the data than the corresponding
logged interaction. This motivates the ideal joint-support-constrained policy
improvement problem
\begin{align}
    \max_{\theta}\quad
    &
    \mathbb{E}_{\substack{
        (s,\tau_e^\beta,\tau_a^\beta)\sim\mathcal{D}\\
        \hat{\tau}_e\sim\pi_\theta(\cdot\mid s)
    }}
    \left[
        Q_{\mathrm{LCB}}(s,\hat{\tau}_e)
    \right]
    \nonumber\\
    \text{s.t.}\quad
    &
    \Delta_J
    \left(
        s,\tau_e^\beta,\hat{\tau}_e,\tau_a^\beta
    \right)
    \leq \epsilon_J,
    \quad \text{a.s.},
    \label{eq:joint_constrained_objective}
\end{align}
where $\epsilon_J\geq 0$ specifies the maximum admissible degradation in joint
interaction support. Equivalently, the constraint requires
\begin{equation}
    p_{\mathcal D}
    (\hat{\tau}_e,\tau_a^\beta\mid s)
    \geq
    e^{-\epsilon_J}
    p_{\mathcal D}
    (\tau_e^\beta,\tau_a^\beta\mid s).
    \label{eq:joint_support_constraint}
\end{equation}

Directly evaluating this constraint is impractical because it requires density
estimation in the high-dimensional joint trajectory space. The key idea of
\textsc{ICDP} is to avoid estimating this joint density directly. Instead, we
show that the joint-support degradation admits an exact decomposition into a
marginal ego-trajectory component and an interaction-specific component. The
latter captures \emph{interaction distribution shift} (IDS), which we estimate
contrastively and use to regularize diffusion-policy optimization.

\subsection{Interaction Distribution Shift}

The joint-support constraint in
(\ref{eq:joint_constrained_objective}) couples two distinct effects:
departure of the candidate ego trajectory from the logged behavior
distribution, and loss of compatibility between that trajectory and the
surrounding-agent evolution observed in the data. Distinguishing these
effects is central to \textsc{ICDP}, since conventional behavior
regularization acts primarily on the former.

Using the conditional factorization
$p_{\mathcal D}(\tau_e,\tau_a\mid s)
= p_{\mathcal D}(\tau_e\mid s)
  p_{\mathcal D}(\tau_a\mid s,\tau_e)$,
the joint-support degradation in
(\ref{eq:joint_support_degradation}) decomposes exactly as
\begin{equation}
    \Delta_J = \Delta_E + \Delta_I,
    \qquad
    \Delta_E =
    \log\frac{p_{\mathcal D}(\tau_e^\beta\mid s)}
              {p_{\mathcal D}(\hat{\tau}_e\mid s)},
    \qquad
    \Delta_I =
    \log\frac{p_{\mathcal D}(\tau_a^\beta\mid s,\tau_e^\beta)}
              {p_{\mathcal D}(\tau_a^\beta\mid s,\hat{\tau}_e)}.
    \label{eq:decomposition}
\end{equation}
Here, $\Delta_E$ captures the ego marginal-support shift and $\Delta_I$
the additional degradation due to ego--agent interaction.
%
%
We refer to $\Delta_I$ as \emph{interaction distribution shift} (IDS).
Unlike $\Delta_E$, which reflects how far the candidate ego trajectory
moves from the logged ego distribution, IDS measures whether the
surrounding-agent evolution observed in the interaction remains
supported when paired with the candidate trajectory. Consequently, a
candidate may remain well supported marginally,
$p_{\mathcal D}(\hat{\tau}_e\mid s)$, yet incur substantial interaction
shift if
$p_{\mathcal D}(\tau_a^\beta\mid s,\hat{\tau}_e)$
is small.

This decomposition isolates the component of distribution shift that is
specific to interaction. Importantly, evaluating IDS does not require
predicting how surrounding agents would respond to the candidate ego
trajectory; it only requires assessing the support of the resulting
ego--agent pairing under the logged interaction distribution. Direct
evaluation of (\ref{eq:decomposition}), however, would require
estimating high-dimensional conditional trajectory densities. We next
show that the required density ratio can instead be recovered
contrastively.

\subsection{Contrastive Recovery of Interaction Distribution Shift}
\label{sec:contrastive_ids}

Our key observation is
that this conditional density need not be modeled explicitly. Instead,
IDS can be recovered by contrasting two notions of support: support of
the complete ego--agent interaction and marginal support of the ego
trajectory alone.

Concretely, we introduce two contrastive scores. A \emph{joint} score
$g_J(s,\tau_e,\tau_a)$ distinguishes logged ego--agent interactions from
mismatched interactions formed using policy-generated ego trajectories,
whereas an \emph{ego} score $g_E(s,\tau_e)$ distinguishes logged ego
trajectories from the same policy-generated candidates irrespective of
the surrounding-agent evolution. Using the same candidate trajectories
in both contrastive problems is important: it ensures that the component
associated with marginal ego-trajectory shift is shared by both scores
and can therefore be eliminated by subtraction. The precise contrastive
distributions and training objectives are given in
Appendix~\ref{app:contrastive_estimation}.

The joint score captures both marginal ego-trajectory shift and
ego--agent dependence, while the ego score captures only the former.
We therefore define the residual interaction score
\begin{equation}
    g_I(s,\tau_e,\tau_a)
    =
    g_J(s,\tau_e,\tau_a)
    -
    g_E(s,\tau_e).
    \label{eq:interaction_score}
\end{equation}
The following result shows that this residual isolates the
interaction-specific component of IDS.

\begin{theorem}[Contrastive recovery of IDS]
\label{thm:interaction_recovery}
Suppose the joint and ego contrastive objectives are constructed using the
same policy-generated candidate trajectories, with balanced class sampling
and common support. Let $g_J^*$ and $g_E^*$ denote their population-optimal
logits, and define the residual interaction score as
\begin{equation}
    g_I^*(s,\tau_e,\tau_a)
    =
    g_J^*(s,\tau_e,\tau_a)
    -
    g_E^*(s,\tau_e).
\end{equation}
Then, for a logged interaction
$(s,\tau_e^\beta,\tau_a^\beta)$ and a candidate ego trajectory
$\hat{\tau}_e$, the difference between their residual interaction scores,
evaluated against the same logged surrounding-agent trajectory
$\tau_a^\beta$, exactly recovers the interaction distribution shift:
\begin{equation}
    \Delta_I
    =
    g_I^*(s,\tau_e^\beta,\tau_a^\beta)
    -
    g_I^*(s,\hat{\tau}_e,\tau_a^\beta).
    \label{eq:ids_recovery}
\end{equation}
\end{theorem}

Theorem~\ref{thm:interaction_recovery} is the central mechanism behind
\textsc{ICDP}. It shows that IDS is identifiable from
the difference of two contrastive scores: the ego score removes the effect
of marginal trajectory novelty from the joint score, leaving precisely the
change in interaction support induced by the candidate ego trajectory.  
In practice, replacing the population
scores with their learned counterparts yields
\begin{equation}
    \widehat{\Delta}_I
    =
    g_I(s,\tau_e^\beta,\tau_a^\beta)
    -
    g_I(s,\hat{\tau}_e,\tau_a^\beta),
    \label{eq:estimated_ids}
\end{equation}
which provides a tractable measure of interaction shift for policy
optimization. The proof of
Theorem~\ref{thm:interaction_recovery} is deferred to
Appendix~\ref{app:contrastive_estimation}.

Beyond its pointwise interpretation, IDS also admits a direct
distributional interpretation. For fixed $s$, $\tau_e^\beta$, and
$\hat{\tau}_e$, averaging the interaction shift over the
surrounding-agent evolution associated with the logged ego trajectory
gives the divergence between the corresponding conditional interaction
distributions.

\begin{corollary}[Distributional interpretation of IDS]
\label{cor:kl_interpretation}
For fixed $s$, $\tau_e^\beta$, and $\hat{\tau}_e$,
\begin{equation}
    \mathbb{E}_{
        \tau_a\sim
        p_{\mathcal D}(\cdot\mid s,\tau_e^\beta)
    }
    \left[
        \Delta_I
    \right]
    =
    D_{\mathrm{KL}}
    \left(
        p_{\mathcal D}(\tau_a\mid s,\tau_e^\beta)
        \,\middle\|\,
        p_{\mathcal D}(\tau_a\mid s,\hat{\tau}_e)
    \right).
    \label{eq:ids_kl}
\end{equation}
\end{corollary}

Thus, while $\Delta_I$ measures interaction shift for an individual
logged ego--agent realization, its expectation quantifies the
distributional change in surrounding-agent behavior associated with
replacing the logged ego trajectory by the candidate. This provides an
information-theoretic interpretation of the interaction constraint used
by \textsc{ICDP}.

\begin{algorithm}[t]
\caption{\textsc{ICDP}}
\label{alg:icdp}
\begin{algorithmic}[1]
\REQUIRE Offline dataset $\mathcal D$
\STATE Pretrain diffusion policy $\pi_\theta$ by behavior cloning
\STATE Initialize critic ensemble $\{Q_{\phi_k}\}_{k=1}^{K}$ and
contrastive estimators $g_J,g_E$
\REPEAT
    \STATE Sample
    $(s,\tau_e^\beta,\tau_a^\beta,R^{(H)},s')\sim\mathcal D$
    \STATE Sample $\hat{\tau}_e\sim\pi_\theta(\cdot\mid s)$
    \STATE Update the critic ensemble
    \STATE Update $g_J$ and $g_E$
    \STATE Compute $\widehat{\Delta}_I$ and the objective in
    \eqref{eq:policy_objective}
    \STATE Update $\pi_\theta$ using NSPG
\UNTIL{convergence}
\RETURN $\pi_\theta$
\end{algorithmic}
\end{algorithm}

\subsection{Interaction-Constrained Policy Optimization (ICDP)}
\label{sec:policy_optimization}

The decomposition in \eqref{eq:decomposition} allows \textsc{ICDP} to
control ego-trajectory shift and interaction distribution shift
separately during policy improvement. For a logged interaction
$(s,\tau_e^\beta,\tau_a^\beta)$ and candidate trajectory
$\hat{\tau}_e\sim\pi_\theta(\cdot\mid s)$, we define the interaction
penalty
\begin{equation}
    C_I
    =
    \left[
        \widehat{\Delta}_I-\epsilon_I
    \right]_+,
    \label{eq:interaction_penalty}
\end{equation}
where $\epsilon_I\geq 0$ denotes the tolerated interaction shift. 
Let
$C_E$ denote the ego-support regularization term, and let
$\mathcal{L}_{\mathrm{BC}}$ denote the behavior-cloning loss on logged
ego trajectories. The policy is optimized according to
\begin{equation}
    \max_\theta
    \;
    \mathbb{E}
    \left[
        Q_{\mathrm{LCB}}(s,\hat{\tau}_e)
        -
        \lambda_E C_E
        -
        \lambda_I C_I
    \right]
    -
    \lambda_{\mathrm{BC}}\mathcal{L}_{\mathrm{BC}},
    \label{eq:policy_objective}
\end{equation}
where $\lambda_E$, $\lambda_I$, and $\lambda_{\mathrm{BC}}$ control
ego-support regularization, interaction-support regularization, and
behavior-cloning regularization, respectively. The BC term anchors the
policy to the demonstrated behavior, while $C_E$ and $C_I$ explicitly
control marginal ego and interaction-level distribution shift.
Since the actor is parameterized as a diffusion model, we optimize
\eqref{eq:policy_objective} using Noisy Space Policy Gradient
(NSPG)~\citep{selimnoisy}. Gradients are propagated through the noisy
denoising process, while the critic and interaction estimator evaluate
the corresponding clean ego trajectory. Algorithm~\ref{alg:icdp} summarizes the complete training procedure.

\section{Experimental Results} \label{sec:experiments}

We evaluate \textsc{ICDP} with three goals: (i) to assess its closed-loop
driving performance relative to prior learning-based planning methods,
(ii) to determine whether interaction-constrained offline RL improves over
behavior cloning and unconstrained policy optimization under a controlled
backbone, and (iii) to study sensitivity to
interaction complexity and regularization strength. Additional results on
InterPlan, implementation details, and extended analyses are provided in the
appendix.

\paragraph{Baselines.}
We compare \textsc{ICDP} against representative prior planners evaluated on
the same nuPlan benchmarks. \textit{PDM-Open}~\citep{dauner2023parting} is the learned
variant of the PDM planning framework; \textit{GameFormer}~\citep{huang2023gameformer}
models interactive prediction and planning using a game-theoretic
transformer architecture; \textit{PlanTF}~\citep{cheng2024rethinking} is an end-to-end
transformer-based motion planner trained from expert demonstrations;
\textit{PLUTO}~\citep{cheng2024pluto} augments imitation-based planning with auxiliary
contrastive objectives; \textit{Diffusion Planner}~\citep{zheng2025diffusion} uses a
diffusion model for multimodal trajectory generation; and
\textit{Flow Planner}~\citep{tan2026flow} adopts flow-based generative modeling for
motion planning. When available, we report variants without rule-based
trajectory refinement to compare the learned policies directly.

\begin{table}[t]
\centering
\caption{
Closed-loop performance on nuPlan Val14, Test14-Hard, and Test14-Random
under non-reactive (NR) and reactive (R) evaluation. The best result in
each column is highlighted in blue.
}
\label{tab:nuplan_prior}

\small
\setlength{\tabcolsep}{13pt}
\renewcommand{\arraystretch}{1.12}

\begin{tabular}{lcccccc}
\toprule
& \multicolumn{2}{c}{Val14}
& \multicolumn{2}{c}{Test14-Hard}
& \multicolumn{2}{c}{Test14-Random} \\
\cmidrule(lr){2-3}
\cmidrule(lr){4-5}
\cmidrule(lr){6-7}
\textbf{Planner}
& \textbf{NR} & \textbf{R}
& \textbf{NR} & \textbf{R}
& \textbf{NR} & \textbf{R} \\
\midrule

PDM-Open
& 53.53 & 54.24
& 33.51 & 35.83
& 52.81 & 57.23 \\

GameFormer w/o refine.
& 13.32 & 8.69
& 7.08 & 6.69
& 11.36 & 9.31 \\

PlanTF
& 84.27 & 76.95
& 69.70 & 61.61
& 85.62 & 79.58 \\

PLUTO w/o refine.
& 88.89 & 78.11
& 70.03 & 59.74
& 89.90 & 78.62 \\

Diffusion Planner
& 89.87 & 82.80
& 75.99 & 69.22
& 89.19 & 82.93 \\

Flow Planner
& 90.43
& 83.31
& 76.47 & 70.42
& 89.88 & 82.93 \\

\midrule

\textsc{ICDP}
& \cellcolor{bestblue} 91.3 & \cellcolor{bestblue} 84.3
& \cellcolor{bestblue}78.34
& \cellcolor{bestblue}72.87
& \cellcolor{bestblue}93.33
& \cellcolor{bestblue}86.34 \\

\bottomrule
\end{tabular}
\end{table}

\subsection{Main results}
\label{sec:experimental_setup}

\paragraph{Closed-loop benchmark performance.}
Table~\ref{tab:nuplan_prior} compares \textsc{ICDP} with prior planners on
the nuPlan closed-loop benchmark. \textsc{ICDP} achieves the strongest
performance on all evaluations. On Test14-Hard, it reaches
$78.34$ in non-reactive evaluation and $72.87$ in reactive evaluation,
exceeding the strongest previously reported results of $76.47$ and $70.42$,
respectively. The improvement is also consistent on Test14-Random, where
\textsc{ICDP} obtains $93.33$ and $86.34$ under non-reactive and reactive
simulation. Notably, the gains are preserved under reactive evaluation,
where surrounding agents respond to the ego vehicle and interaction quality
becomes particularly important.

\paragraph{Controlled policy improvement.}
We next isolate the effect of reward-driven policy optimization and
interaction-aware regularization using the same policy backbone and training
data. Table~\ref{tab:controlled_rl} reports performance across ten evaluation
checkpoints spaced by 2k optimization steps. Offline RL improves reactive
performance over BC on both Test14-Hard and Test14-Random, but its gains are
not consistent across the non-reactive evaluations. Incorporating the
interaction constraint yields the strongest performance in five of the six
settings. In particular, \textsc{ICDP} improves the reactive score from
$70.07$ to $72.87$ on Test14-Hard and from $83.02$ to $86.34$ on
Test14-Random relative to unconstrained Offline RL. These results indicate
that interaction-aware regularization is particularly beneficial when the
evaluation explicitly captures the response of surrounding traffic.

\begin{table}[t]
\centering
\caption{
Controlled comparison of BC, unconstrained Offline RL, and \textsc{ICDP}.
Results are averaged over ten evaluation checkpoints spaced by 2k
optimization steps and reported as mean $\pm$ std.
}
\label{tab:controlled_rl}

\small
\setlength{\tabcolsep}{4pt}
\renewcommand{\arraystretch}{1.12}

\begin{tabular}{lcccccc}
\toprule
& \multicolumn{2}{c}{Val14}
& \multicolumn{2}{c}{Test14-Hard}
& \multicolumn{2}{c}{Test14-Random} \\
\cmidrule(lr){2-3}
\cmidrule(lr){4-5}
\cmidrule(lr){6-7}
\textbf{Method}
& \textbf{NR} & \textbf{R}
& \textbf{NR} & \textbf{R}
& \textbf{NR} & \textbf{R} \\
\midrule

BC
& $89.8 \pm 1.2$ & $82.4 \pm 1.09$
& $78.42 \pm 0.69$
& $68.21 \pm 0.95$
& $91.98 \pm 0.62$
& $82.71 \pm 0.53$ \\

Offline RL
& $89.4 \pm 1.74$ & $83.4 \pm 2.11$
& \cellcolor{bestblue}$78.75 \pm 0.75$
& $70.07 \pm 1.08$
& $90.74 \pm 0.70$
& $83.02 \pm 1.08$ \\

\textsc{ICDP}
& \cellcolor{bestblue} $91.3 \pm 1.91$ & \cellcolor{bestblue} $84.3 \pm 2.04$
& $78.34 \pm 1.03$
& \cellcolor{bestblue}$72.87 \pm 2.01$
& \cellcolor{bestblue}$93.33 \pm 0.85$
& \cellcolor{bestblue}$86.34 \pm 1.18$ \\

\bottomrule
\end{tabular}
\end{table}

\subsection{Analysis of Interaction-Aware Regularization}
\label{sec:interaction_analysis}

We analyze how the interaction-aware component of \textsc{ICDP} depends on
the strength of ego- and interaction-support regularization and on the amount
of surrounding-agent context provided to the support estimator. 
For compactness, we report the
average closed-loop score across Test14-Hard and Test14-Random under both
non-reactive and reactive evaluation.

\begin{figure*}[t]
    \centering

    \begin{minipage}[t]{0.32\textwidth}
        \centering
        \includegraphics[width=\linewidth]
        {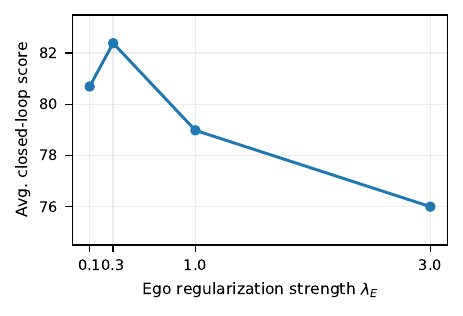}
        
        \small (a) Ego-support weight $\lambda_E$
    \end{minipage}
    \hfill
    \begin{minipage}[t]{0.32\textwidth}
        \centering
        \includegraphics[width=\linewidth]
        {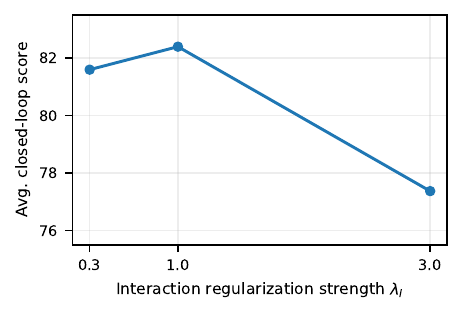}
        
        \small (b) Interaction-support weight $\lambda_I$
    \end{minipage}
    \hfill
    \begin{minipage}[t]{0.32\textwidth}
        \centering
        \includegraphics[width=\linewidth]
        {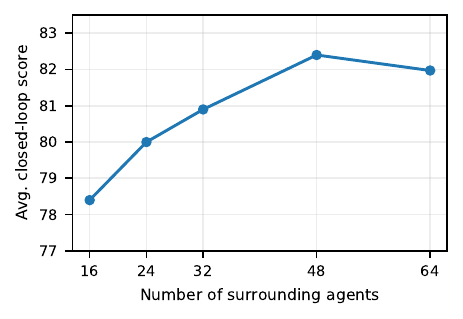}
        
        \small (c) Number of surrounding agents
    \end{minipage}

    \caption{
    Sensitivity of \textsc{ICDP} to
    (a) ego-support regularization,
    (b) interaction-support regularization, and
    (c) the number of surrounding agents.
    Scores are averaged across Test14-Hard and Test14-Random under
    non-reactive and reactive evaluation.
    }
    \label{fig:interaction_sensitivity}
\end{figure*}

\paragraph{Regularization sensitivity.}
Figure~\ref{fig:interaction_sensitivity}(a--b) examines the effect of the
ego- and interaction-support regularization strengths. For ego-support
regularization, performance improves from $\lambda_E=0.1$ to $0.3$, but
decreases as the weight is increased further, indicating that moderate
regularization provides the best balance between preserving data support
and allowing reward-driven policy improvement. A similar pattern is observed
for interaction regularization: $\lambda_I=1$ performs better than both
weaker and stronger regularization among the tested values. We therefore use
$\lambda_E=0.3$ and $\lambda_I=1$ in the reported \textsc{ICDP}
configuration.

\paragraph{Number of surrounding agents.}
Figure~\ref{fig:interaction_sensitivity}(c) studies the amount of
multi-agent context used by the interaction-support estimator. Performance
improves consistently as the number of surrounding agents increases from
$16$ to $48$. Expanding the context further to $64$ agents yields a small
decrease in average performance, from $82.40$ to $81.97$. These results
suggest that incorporating a sufficiently broad set of surrounding agents is
beneficial, while increasing the interaction context beyond this point
provides limited additional benefit. 

\subsection{Real-World Deployment}
\label{sec:real_world}

Finally, we deploy \textsc{ICDP} on a full-scale autonomous truck to assess
its behavior outside simulation. The experiments were conducted on a
snow-covered test track and covered two complementary settings:
multi-vehicle interactions and standalone driving on curved roads.



\begin{figure*}[t]
    \centering

    \begin{minipage}[t]{0.24\textwidth}
        \centering
        \includegraphics[
            width=\linewidth,
            height=3.0cm
        ]{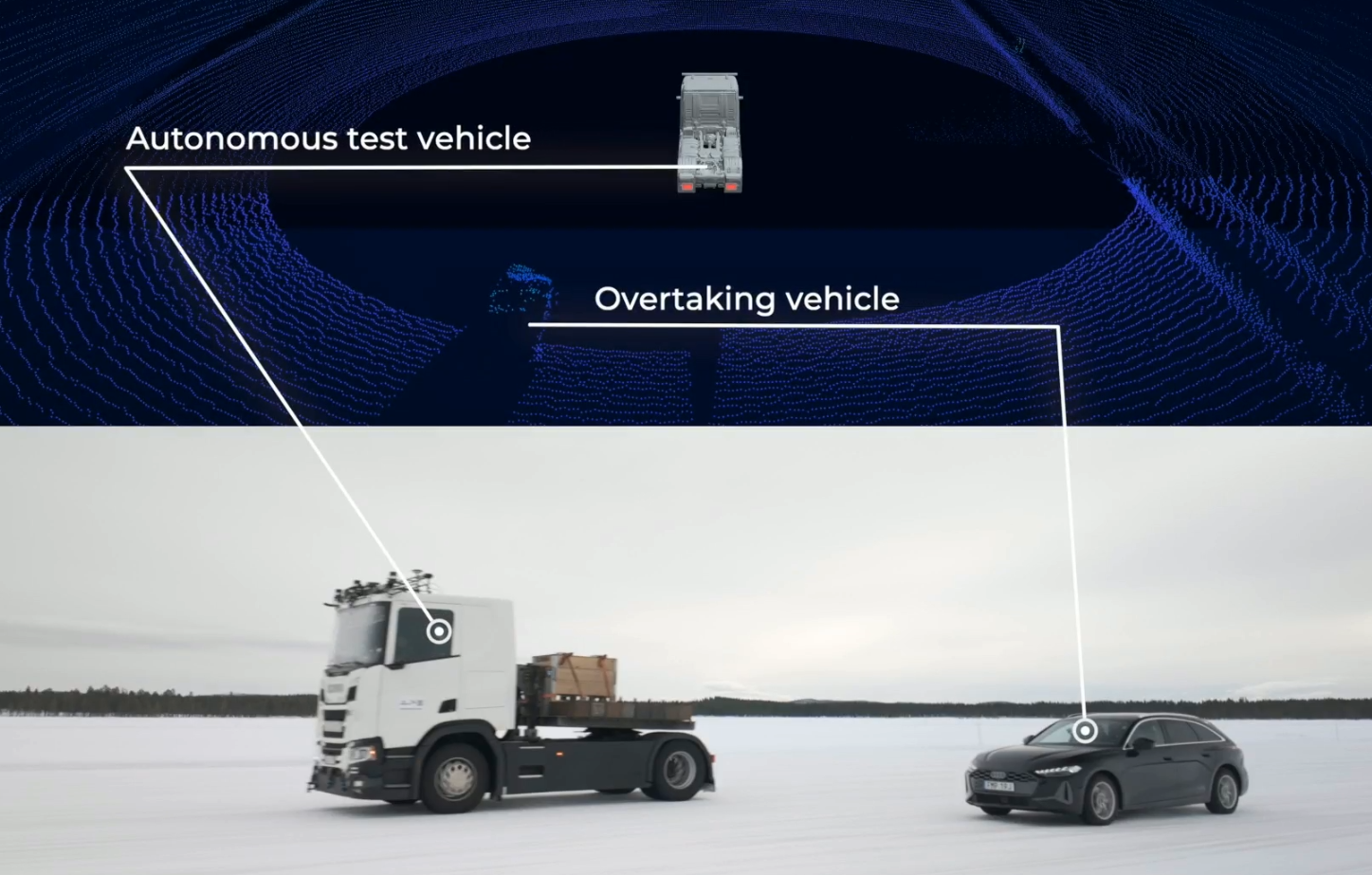}
        \small (a)
    \end{minipage}
    \hfill
    \begin{minipage}[t]{0.24\textwidth}
        \centering
        \includegraphics[
            width=\linewidth,
            height=3.0cm
        ]{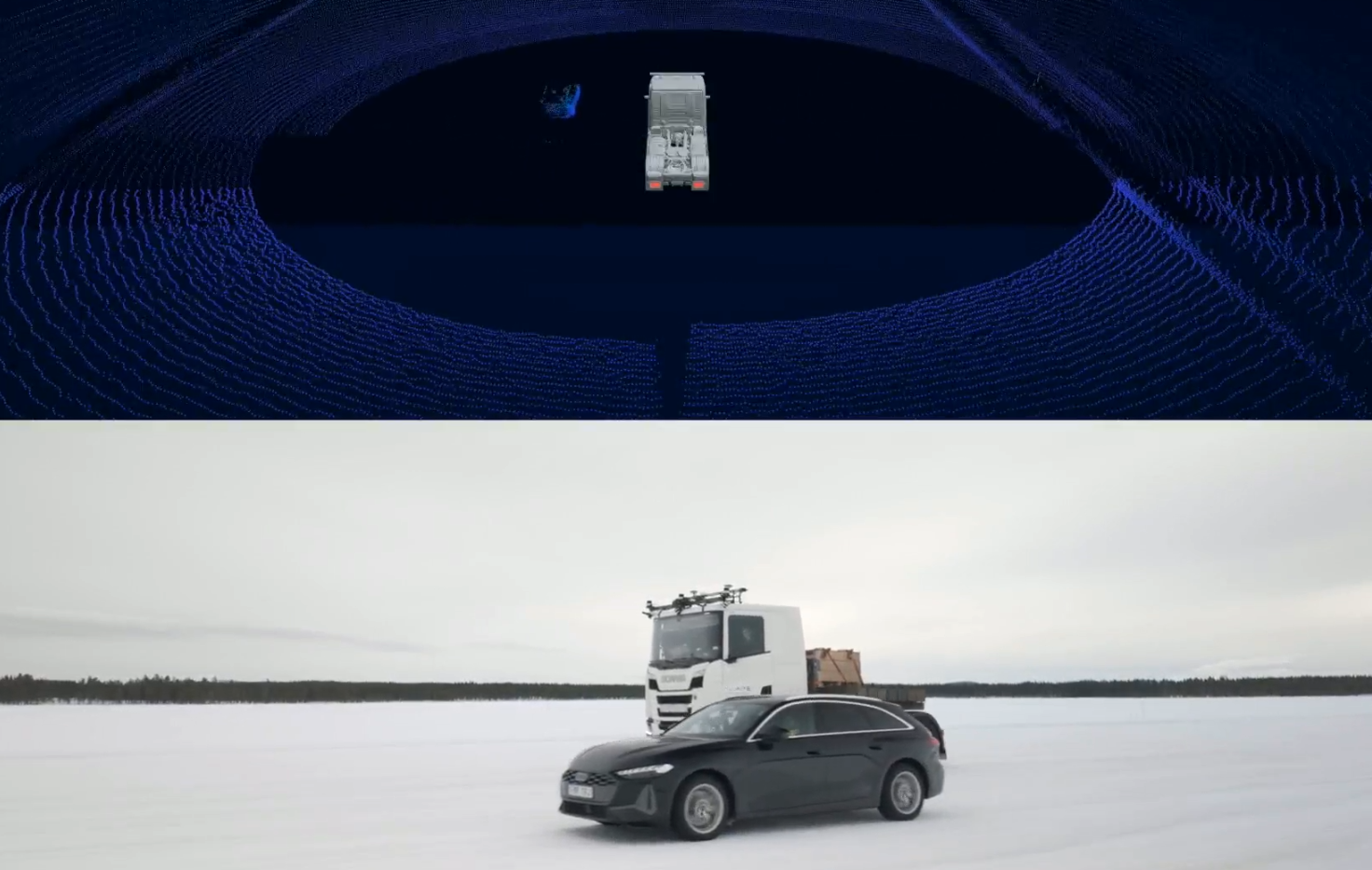}
        \small (b)
    \end{minipage}
    \hfill
    \begin{minipage}[t]{0.24\textwidth}
        \centering
        \includegraphics[
            width=\linewidth,
            height=3.0cm
        ]{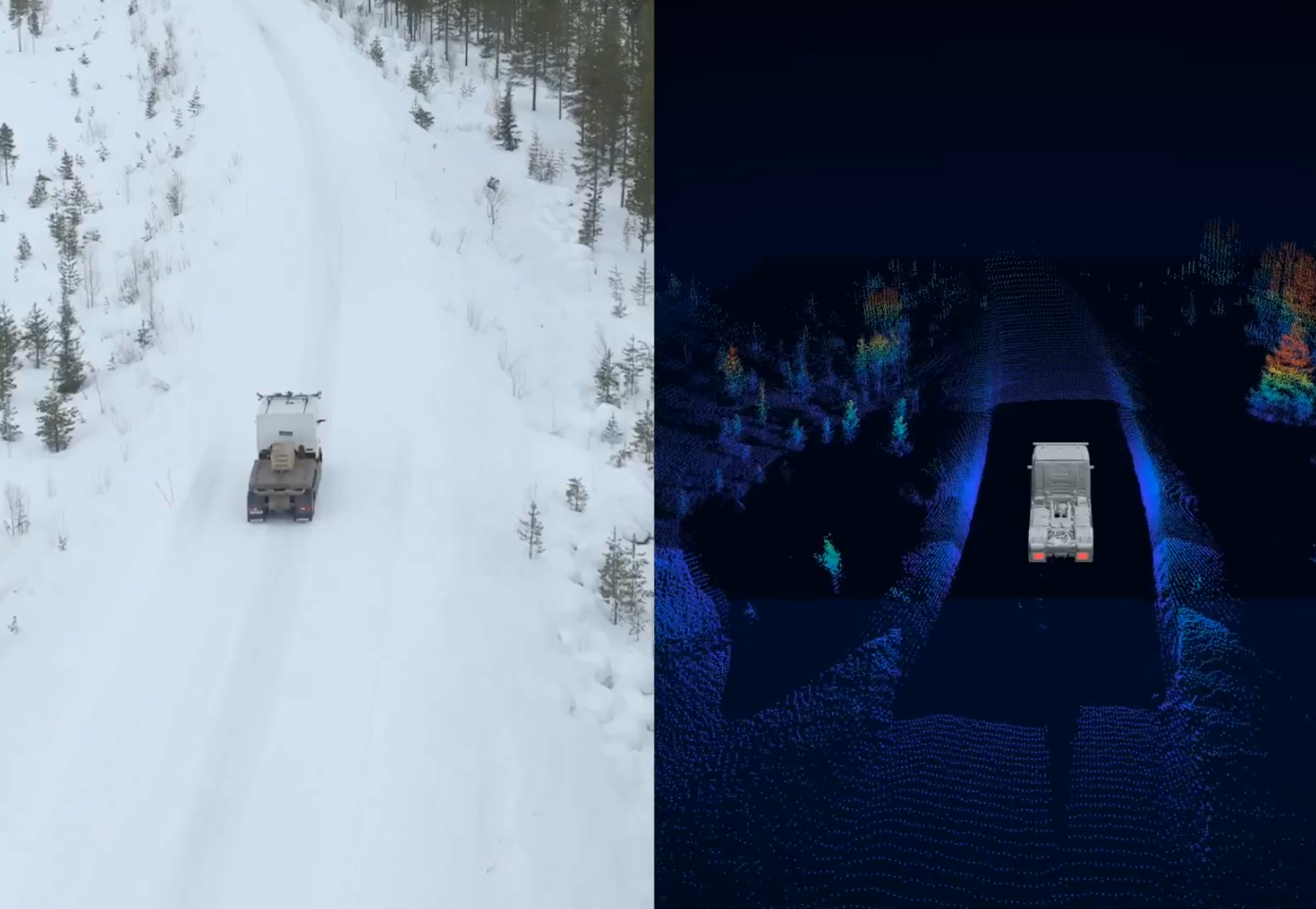}
        \small (c)
    \end{minipage}
    \hfill
    \begin{minipage}[t]{0.24\textwidth}
        \centering
        \includegraphics[
            width=\linewidth,
            height=3.0cm
        ]{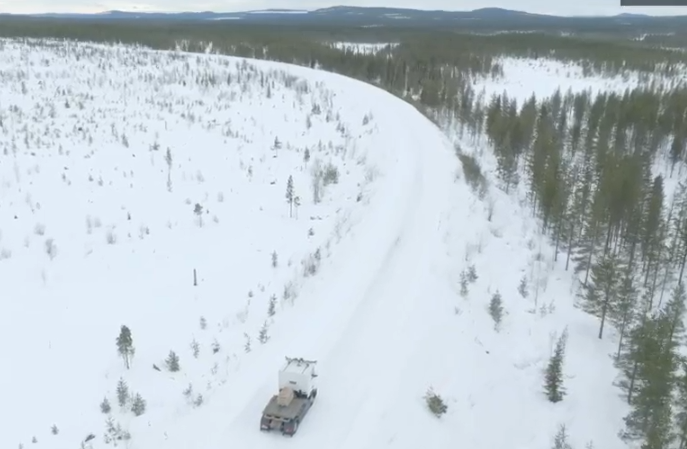}
        \small (d)
    \end{minipage}

    \caption{
    Real-world deployment of \textsc{ICDP} on a full-scale autonomous
    truck. (a--b) Representative multi-agent interaction trial involving
    an overtaking vehicle, shown from external and onboard-perception
    viewpoints. (c--d) Standalone driving on curved road segments under
    real sensing and vehicle dynamics. 
    }\vspace{-3mm}
    \label{fig:real_world}
\end{figure*}

\paragraph{Interactive driving.}
We evaluate \textsc{ICDP} in real-world multi-vehicle scenarios involving
overtaking and lead-vehicle following under varying traffic configurations.
Figure~\ref{fig:real_world}(a--b) shows representative overtaking examples
together with the corresponding onboard perception view. These experiments
provide qualitative evidence that the policy can execute interaction-aware
maneuvers under real sensing and vehicle dynamics.

\paragraph{Standalone driving.}
We additionally evaluate the policy on curved road segments without nearby
interacting vehicles. Representative examples are shown in
Fig.~\ref{fig:real_world}(c--d), illustrating nominal trajectory generation
and execution under real road geometry, sensing, and vehicle dynamics.

Overall, the real-world experiments provide qualitative evidence that
\textsc{ICDP} can be deployed across both interactive and nominal driving
conditions.

\section{Conclusion and Future Work}
\label{sec:conclusion}

We introduced \textsc{ICDP}, an offline RL framework
for diffusion-based driving that explicitly controls distribution shift at
the level of multi-agent interactions. By decomposing joint-support
degradation into ego-trajectory and interaction-specific components, we
formalized \emph{interaction distribution shift} (IDS) and showed that it
can be recovered contrastively without explicit conditional density
estimation or prediction of surrounding-agent responses. \textsc{ICDP}
combines this interaction constraint with chunk-level value learning and
diffusion-policy optimization to improve long-horizon driving performance
while preserving support under the logged interaction distribution.

\update{
Closed-loop evaluations on nuPlan show that ICDP consistently
outperforms prior planners across non-reactive and
reactive settings. Controlled comparisons further support the benefit
of interaction-aware regularization, with particularly clear gains over both
behavior cloning and unconstrained offline RL across reactive
evaluations. Real-world deployment provides qualitative evidence
of the framework's feasibility on a full-scale autonomous truck. A natural next step is to investigate online fine-tuning of the
offline-trained \textsc{ICDP} policy, using closed-loop feedback
to further improve driving performance and adapt to interactions
insufficiently represented in the logged data.
}

\newpage
\bibliography{ref} 

@article{fan2018baidu,
  title={Baidu apollo em motion planner},
  author={Fan, Haoyang and Zhu, Fan and Liu, Changchun and Zhang, Liangliang and Zhuang, Li and Li, Dong and Zhu, Weicheng and Hu, Jiangtao and Li, Hongye and Kong, Qi},
  journal={arXiv preprint arXiv:1807.08048},
  year={2018}
}

@inproceedings{hawke2020urban,
  title={Urban driving with conditional imitation learning},
  author={Hawke, Jeffrey and Shen, Richard and Gurau, Corina and Sharma, Siddharth and Reda, Daniele and Nikolov, Nikolay and Mazur, Przemys{\l}aw and Micklethwaite, Sean and Griffiths, Nicolas and Shah, Amar and others},
  booktitle={2020 IEEE International Conference on Robotics and Automation (ICRA)},
  pages={251--257},
  year={2020},
  organization={IEEE}
}

@article{bojarski2016end,
  title={End to end learning for self-driving cars},
  author={Bojarski, Mariusz and Del Testa, Davide and Dworakowski, Daniel and Firner, Bernhard and Flepp, Beat and Goyal, Prasoon and Jackel, Lawrence D and Monfort, Mathew and Muller, Urs and Zhang, Jiakai and others},
  journal={arXiv preprint arXiv:1604.07316},
  year={2016}
}

@article{chitta2022transfuser,
  title={Transfuser: Imitation with transformer-based sensor fusion for autonomous driving},
  author={Chitta, Kashyap and Prakash, Aditya and Jaeger, Bernhard and Yu, Zehao and Renz, Katrin and Geiger, Andreas},
  journal={IEEE transactions on pattern analysis and machine intelligence},
  volume={45},
  number={11},
  pages={12878--12895},
  year={2022},
  publisher={IEEE}
}

@inproceedings{cheng2024rethinking,
  title={Rethinking imitation-based planners for autonomous driving},
  author={Cheng, Jie and Chen, Yingbing and Mei, Xiaodong and Yang, Bowen and Li, Bo and Liu, Ming},
  booktitle={2024 IEEE International Conference on Robotics and Automation (ICRA)},
  pages={14123--14130},
  year={2024},
  organization={IEEE}
}

@article{zheng2025diffusion,
  title={Diffusion-based planning for autonomous driving with flexible guidance},
  author={Zheng, Yinan and Liang, Ruiming and Zheng, Kexin and Zheng, Jinliang and Mao, Liyuan and Li, Jianxiong and Gu, Weihao and Ai, Rui and Li, Shengbo Eben and Zhan, Xianyuan and others},
  journal={arXiv preprint arXiv:2501.15564},
  year={2025}
}

@inproceedings{huang2023gameformer,
  title={Gameformer: Game-theoretic modeling and learning of transformer-based interactive prediction and planning for autonomous driving},
  author={Huang, Zhiyu and Liu, Haochen and Lv, Chen},
  booktitle={Proceedings of the IEEE/CVF International Conference on Computer Vision},
  pages={3903--3913},
  year={2023}
}

@article{cheng2024pluto,
  title={Pluto: Pushing the limit of imitation learning-based planning for autonomous driving},
  author={Cheng, Jie and Chen, Yingbing and Chen, Qifeng},
  journal={arXiv preprint arXiv:2404.14327},
  year={2024}
}

@article{bansal2018chauffeurnet,
  title={Chauffeurnet: Learning to drive by imitating the best and synthesizing the worst},
  author={Bansal, Mayank and Krizhevsky, Alex and Ogale, Abhijit},
  journal={arXiv preprint arXiv:1812.03079},
  year={2018}
}

@inproceedings{dauner2023parting,
  title={Parting with misconceptions about learning-based vehicle motion planning},
  author={Dauner, Daniel and Hallgarten, Marcel and Geiger, Andreas and Chitta, Kashyap},
  booktitle={Conference on Robot Learning},
  pages={1268--1281},
  year={2023},
  organization={PMLR}
}

@article{levine2020offline,
  title={Offline reinforcement learning: Tutorial, review, and perspectives on open problems},
  author={Levine, Sergey and Kumar, Aviral and Tucker, George and Fu, Justin},
  journal={arXiv preprint arXiv:2005.01643},
  year={2020}
}

@article{kumar2020conservative,
  title={Conservative q-learning for offline reinforcement learning},
  author={Kumar, Aviral and Zhou, Aurick and Tucker, George and Levine, Sergey},
  journal={Advances in neural information processing systems},
  volume={33},
  pages={1179--1191},
  year={2020}
}

@article{kostrikov2021offline,
  title={Offline reinforcement learning with implicit q-learning},
  author={Kostrikov, Ilya and Nair, Ashvin and Levine, Sergey},
  journal={arXiv preprint arXiv:2110.06169},
  year={2021}
}

@article{peng2019advantage,
  title={Advantage-weighted regression: Simple and scalable off-policy reinforcement learning},
  author={Peng, Xue Bin and Kumar, Aviral and Zhang, Grace and Levine, Sergey},
  journal={arXiv preprint arXiv:1910.00177},
  year={2019}
}

@inproceedings{casas2021mp3,
  title={Mp3: A unified model to map, perceive, predict and plan},
  author={Casas, Sergio and Sadat, Abbas and Urtasun, Raquel},
  booktitle={Proceedings of the IEEE/CVF Conference on Computer Vision and Pattern Recognition},
  pages={14403--14412},
  year={2021}
}

@inproceedings{codevilla2019exploring,
  title={Exploring the limitations of behavior cloning for autonomous driving},
  author={Codevilla, Felipe and Santana, Eder and L{\'o}pez, Antonio M and Gaidon, Adrien},
  booktitle={Proceedings of the IEEE/CVF international conference on computer vision},
  pages={9329--9338},
  year={2019}
}

@inproceedings{liang2020learning,
  title={Learning lane graph representations for motion forecasting},
  author={Liang, Ming and Yang, Bin and Hu, Rui and Chen, Yun and Liao, Renjie and Feng, Song and Urtasun, Raquel},
  booktitle={European Conference on Computer Vision},
  pages={541--556},
  year={2020},
  organization={Springer}
}

@inproceedings{zhang2025carplanner,
  title={Carplanner: Consistent auto-regressive trajectory planning for large-scale reinforcement learning in autonomous driving},
  author={Zhang, Dongkun and Liang, Jiaming and Guo, Ke and Lu, Sha and Wang, Qi and Xiong, Rong and Miao, Zhenwei and Wang, Yue},
  booktitle={Proceedings of the Computer Vision and Pattern Recognition Conference},
  pages={17239--17248},
  year={2025}
}

@inproceedings{kendall2019learning,
  title={Learning to drive in a day},
  author={Kendall, Alex and Hawke, Jeffrey and Janz, David and Mazur, Przemyslaw and Reda, Daniele and Allen, John-Mark and Lam, Vinh-Dieu and Bewley, Alex and Shah, Amar},
  booktitle={2019 international conference on robotics and automation (ICRA)},
  pages={8248--8254},
  year={2019},
  organization={IEEE}
}

@article{yang2026raw2drive,
  title={Raw2drive: Reinforcement learning with aligned world models for end-to-end autonomous driving (in carla v2)},
  author={Yang, Zhenjie and Jia, Xiaosong and Li, Qifeng and Yang, Xue and Yao, Maoqing and Yan, Junchi},
  journal={Advances in Neural Information Processing Systems},
  volume={38},
  pages={134122--134147},
  year={2026}
}

@inproceedings{toromanoff2020end,
  title={End-to-end model-free reinforcement learning for urban driving using implicit affordances},
  author={Toromanoff, Marin and Wirbel, Emilie and Moutarde, Fabien},
  booktitle={Proceedings of the IEEE/CVF conference on computer vision and pattern recognition},
  pages={7153--7162},
  year={2020}
}

@article{jaeger2025carl,
  title={Carl: Learning scalable planning policies with simple rewards},
  author={Jaeger, Bernhard and Dauner, Daniel and Bei{\ss}wenger, Jens and Gerstenecker, Simon and Chitta, Kashyap and Geiger, Andreas},
  journal={arXiv preprint arXiv:2504.17838},
  year={2025}
}

@article{wu2023iplan,
  title={iplan: Intent-aware planning in heterogeneous traffic via distributed multi-agent reinforcement learning},
  author={Wu, Xiyang and Chandra, Rohan and Guan, Tianrui and Bedi, Amrit Singh and Manocha, Dinesh},
  journal={arXiv preprint arXiv:2306.06236},
  year={2023}
}

@article{wang2022diffusion,
  title={Diffusion policies as an expressive policy class for offline reinforcement learning},
  author={Wang, Zhendong and Hunt, Jonathan J and Zhou, Mingyuan},
  journal={arXiv preprint arXiv:2208.06193},
  year={2022}
}

@inproceedings{fang2025diffusion,
  title={Diffusion actor-critic: Formulating constrained policy iteration as diffusion noise regression for offline reinforcement learning},
  author={Fang, Linjiajie and Liu, Ruoxue and Zhang, Jing and Wang, Wenjia and Jing, Bingyi},
  booktitle={International Conference on Learning Representations},
  volume={2025},
  pages={75743--75763},
  year={2025}
}

@article{diehl2021umbrella,
  title={Umbrella: Uncertainty-aware model-based offline reinforcement learning leveraging planning},
  author={Diehl, Christopher and Sievernich, Timo and Kr{\"u}ger, Martin and Hoffmann, Frank and Bertram, Torsten},
  journal={arXiv preprint arXiv:2111.11097},
  year={2021}
}

@article{rowe2024ctrl,
  title={Ctrl-sim: Reactive and controllable driving agents with offline reinforcement learning},
  author={Rowe, Luke and Girgis, Roger and Gosselin, Anthony and Carrez, Bruno and Golemo, Florian and Heide, Felix and Paull, Liam and Pal, Christopher},
  journal={arXiv preprint arXiv:2403.19918},
  year={2024}
}

@article{li2023uncertainty,
  title={Uncertainty-aware decision transformer for stochastic driving environments},
  author={Li, Zenan and Nie, Fan and Sun, Qiao and Da, Fang and Zhao, Hang},
  journal={arXiv preprint arXiv:2309.16397},
  year={2023}
}

@article{li2026reinforcement,
  title={Reinforcement learning with action chunking},
  author={Li, Qiyang and Zhou, Zhiyuan Paul and Levine, Sergey},
  journal={Advances in Neural Information Processing Systems},
  volume={38},
  pages={55518--55553},
  year={2026}
}

@inproceedings{wang2025adawm,
  title={Adawm: Adaptive world model based planning for autonomous driving},
  author={Wang, Hang and Ye, Xin and Tao, Feng and Pan, Chenbin and Mallik, Abhirup and Yaman, Burhan and Ren, Liu and Zhang, Junshan},
  booktitle={International Conference on Learning Representations},
  volume={2025},
  pages={85591--85615},
  year={2025}
}

@inproceedings{noguchi2026pseudo,
  title={Pseudo-Expert Regularized Offline RL for End-to-End Autonomous Driving in Photorealistic Closed-Loop Environments},
  author={Noguchi, Chihiro and Yamamoto, Takaki},
  booktitle={Proceedings of the IEEE/CVF Conference on Computer Vision and Pattern Recognition},
  pages={1096--1105},
  year={2026}
}

@article{huang2025lane,
  title={Lane-changing policy offline reinforcement learning of autonomous vehicles based on BEAR algorithm with support set constraints},
  author={Huang, Caixia and Wang, Yuxiang and Zhang, Zhiyong and Feng, Wenming and Huang, Dayang},
  journal={Proceedings of the Institution of Mechanical Engineers, Part D: Journal of Automobile Engineering},
  volume={239},
  number={9},
  pages={3784--3797},
  year={2025},
  publisher={SAGE Publications Sage UK: London, England}
}

@inproceedings{hu2023planning,
  title={Planning-oriented autonomous driving},
  author={Hu, Yihan and Yang, Jiazhi and Chen, Li and Li, Keyu and Sima, Chonghao and Zhu, Xizhou and Chai, Siqi and Du, Senyao and Lin, Tianwei and Wang, Wenhai and others},
  booktitle={2023 IEEE/CVF Conference on Computer Vision and Pattern Recognition (CVPR)},
  pages={17853--17862},
  year={2023},
  organization={IEEE}
}

@inproceedings{jiang2023vad,
  title={Vad: Vectorized scene representation for efficient autonomous driving},
  author={Jiang, Bo and Chen, Shaoyu and Xu, Qing and Liao, Bencheng and Chen, Jiajie and Zhou, Helong and Zhang, Qian and Liu, Wenyu and Huang, Chang and Wang, Xinggang},
  booktitle={2023 IEEE/CVF International Conference on Computer Vision (ICCV)},
  pages={8306--8316},
  year={2023},
  organization={IEEE}
}

@inproceedings{liao2025diffusiondrive,
  title={Diffusiondrive: Truncated diffusion model for end-to-end autonomous driving},
  author={Liao, Bencheng and Chen, Shaoyu and Yin, Haoran and Jiang, Bo and Wang, Cheng and Yan, Sixu and Zhang, Xinbang and Li, Xiangyu and Zhang, Ying and Zhang, Qian and others},
  booktitle={2025 IEEE/CVF Conference on Computer Vision and Pattern Recognition (CVPR)},
  pages={12037--12047},
  year={2025},
  organization={IEEE}
}

@inproceedings{li2024think2drive,
  title={Think2drive: Efficient reinforcement learning by thinking with latent world model for autonomous driving (in carla-v2)},
  author={Li, Qifeng and Jia, Xiaosong and Wang, Shaobo and Yan, Junchi},
  booktitle={European conference on computer vision},
  pages={142--158},
  year={2024},
  organization={Springer}
}

@article{kumar2019stabilizing,
  title={Stabilizing off-policy q-learning via bootstrapping error reduction},
  author={Kumar, Aviral and Fu, Justin and Soh, Matthew and Tucker, George and Levine, Sergey},
  journal={Advances in neural information processing systems},
  volume={32},
  year={2019}
}

@article{fujimoto2021minimalist,
  title={A minimalist approach to offline reinforcement learning},
  author={Fujimoto, Scott and Gu, Shixiang Shane},
  journal={Advances in neural information processing systems},
  volume={34},
  pages={20132--20145},
  year={2021}
}

@article{gao2025behavior,
  title={Behavior-regularized diffusion policy optimization for offline reinforcement learning},
  author={Gao, Chen-Xiao and Wu, Chenyang and Cao, Mingjun and Xiao, Chenjun and Yu, Yang and Zhang, Zongzhang},
  journal={arXiv preprint arXiv:2502.04778},
  year={2025}
}

@inproceedings{
selimnoisy,
title={Noisy-Space Policy Gradient for Diffusion Policies in Offline Reinforcement Learning},
author={Mahmoud Selim and Cristina Cipriani and Karl Henrik Johansson},
booktitle={Forty-third International Conference on Machine Learning},
year={2026},
}

@article{nguyen2025comadice,
  title={Comadice: Offline cooperative multi-agent reinforcement learning with stationary distribution shift regularization},
  author={Bui, The Viet and Nguyen, Thanh Hong and Mai, Tien},
  journal={arXiv preprint arXiv:2410.01954},
  year={2024}
}

@article{tan2026flow,
  title={Flow matching-based autonomous driving planning with advanced interactive behavior modeling},
  author={Tan, Tianyi and Zheng, Yinan and Liang, Ruiming and Wang, Zexu and Zheng, Kexin and Zheng, Jinliang and Li, Jianxiong and Zhan, Xianyuan and Liu, Jingjing},
  journal={Advances in Neural Information Processing Systems},
  volume={38},
  pages={38310--38335},
  year={2026}
}
\bibliographystyle{iclr2027_conference}

\newpage

\appendix
\section{Contrastive Estimation of Interaction Distribution Shift}
\label{app:contrastive_estimation}

This section provides the detailed construction of the contrastive
objectives used to estimate interaction distribution shift (IDS), together
with the proof of Theorem~\ref{thm:interaction_recovery} and the
information-theoretic interpretation of $\Delta_I$.

\subsection{Contrastive Objectives}

Let $\nu(\tau_e\mid s)$ denote the distribution from which candidate ego
trajectories are sampled during policy optimization. The same candidate
distribution is used in both contrastive objectives described below.

\paragraph{Joint interaction classification.}
The first contrastive objective distinguishes ego--agent interactions
observed in the offline dataset from interactions formed by pairing a
policy-generated ego trajectory with the surrounding-agent trajectory
observed in the same scene. Its positive distribution is
\begin{equation}
    p_J^{+}(s,\tau_e,\tau_a)
    =
    d_{\mathcal D}(s)\,
    p_{\mathcal D}(\tau_e,\tau_a\mid s),
    \label{eq:app_joint_positive}
\end{equation}
where $d_{\mathcal D}(s)$ denotes the empirical state distribution.
The corresponding negative distribution is
\begin{equation}
    p_J^{-}(s,\tau_e,\tau_a)
    =
    d_{\mathcal D}(s)\,
    \nu(\tau_e\mid s)\,
    p_{\mathcal D}(\tau_a\mid s).
    \label{eq:app_joint_negative}
\end{equation}
\update{
Thus, the negative construction combines the candidate ego marginal
$\nu(\tau_e \mid s)$ with the logged surrounding-agent marginal
$p_{\mathcal{D}}(\tau_a \mid s)$. The factorization in Eq.~\ref{eq:app_joint_negative} follows
from sampling these trajectories independently conditional on the scene.
This independence is a property of the negative-sampling construction;
the contrastive recovery of IDS established below does not require
conditional independence between ego and surrounding-agent trajectories
under the logged joint distribution
$p_{\mathcal{D}}(\tau_e, \tau_a \mid s)$.
}

Operationally, given a logged sample
$(s,\tau_e^\beta,\tau_a^\beta)\sim\mathcal D$ and a candidate trajectory
$\hat{\tau}_e\sim\nu(\cdot\mid s)$, the positive and negative examples
are respectively
\begin{equation}
    (s,\tau_e^\beta,\tau_a^\beta),
    \qquad
    (s,\hat{\tau}_e,\tau_a^\beta).
\end{equation}

Let $g_J(s,\tau_e,\tau_a)$ denote the logit of the corresponding binary
classifier.

\paragraph{Marginal ego classification.}
The joint classifier is sensitive both to whether the ego trajectory is
supported by the logged behavior distribution and to whether it is
compatible with the surrounding-agent trajectory. To separate these two
effects, we introduce a second contrastive objective defined only over ego
trajectories. Its positive and negative distributions are
\begin{align}
    p_E^{+}(s,\tau_e)
    &=
    d_{\mathcal D}(s)\,
    p_{\mathcal D}(\tau_e\mid s),
    \label{eq:app_ego_positive}\\
    p_E^{-}(s,\tau_e)
    &=
    d_{\mathcal D}(s)\,
    \nu(\tau_e\mid s).
    \label{eq:app_ego_negative}
\end{align}
Let $g_E(s,\tau_e)$ denote the corresponding classifier logit. In
practice, the negative ego example is the same candidate trajectory
$\hat{\tau}_e$ used to construct the negative joint interaction.

Under balanced positive and negative sampling, the two classifiers are
trained using the binary cross-entropy objectives
\begin{align}
    \mathcal L_J
    &=
    -
    \mathbb E_{p_J^{+}}
    \left[
        \log \sigma(g_J)
    \right]
    -
    \mathbb E_{p_J^{-}}
    \left[
        \log (1-\sigma(g_J))
    \right],
    \label{eq:app_joint_loss}\\
    \mathcal L_E
    &=
    -
    \mathbb E_{p_E^{+}}
    \left[
        \log \sigma(g_E)
    \right]
    -
    \mathbb E_{p_E^{-}}
    \left[
        \log (1-\sigma(g_E))
    \right],
    \label{eq:app_ego_loss}
\end{align}
where $\sigma(\cdot)$ denotes the logistic sigmoid.

As defined in the main text, the residual interaction score is
\begin{equation}
    g_I(s,\tau_e,\tau_a)
    =
    g_J(s,\tau_e,\tau_a)
    -
    g_E(s,\tau_e).
    \label{eq:app_interaction_score}
\end{equation}

\subsection{Proof of Theorem~\ref{thm:interaction_recovery}}

For balanced binary classification, the population-optimal logit is the
log-density ratio between the positive and negative distributions.
Applying this result to
\eqref{eq:app_joint_positive}--\eqref{eq:app_joint_negative} gives
\begin{align}
    g_J^*(s,\tau_e,\tau_a)
    &=
    \log
    \frac{
        p_J^{+}(s,\tau_e,\tau_a)
    }{
        p_J^{-}(s,\tau_e,\tau_a)
    }
    \nonumber\\
    &=
    \log
    \frac{
        p_{\mathcal D}(\tau_e,\tau_a\mid s)
    }{
        \nu(\tau_e\mid s)\,
        p_{\mathcal D}(\tau_a\mid s)
    }.
    \label{eq:app_optimal_joint}
\end{align}
Similarly, the population-optimal ego classifier satisfies
\begin{align}
    g_E^*(s,\tau_e)
    &=
    \log
    \frac{
        p_E^{+}(s,\tau_e)
    }{
        p_E^{-}(s,\tau_e)
    }
    \nonumber\\
    &=
    \log
    \frac{
        p_{\mathcal D}(\tau_e\mid s)
    }{
        \nu(\tau_e\mid s)
    }.
    \label{eq:app_optimal_ego}
\end{align}

Subtracting \eqref{eq:app_optimal_ego} from
\eqref{eq:app_optimal_joint}, the candidate-trajectory distribution
$\nu(\tau_e\mid s)$ cancels:
\begin{align}
    g_I^*(s,\tau_e,\tau_a)
    &=
    g_J^*(s,\tau_e,\tau_a)
    -
    g_E^*(s,\tau_e)
    \nonumber\\
    &=
    \log
    \frac{
        p_{\mathcal D}(\tau_e,\tau_a\mid s)
    }{
        p_{\mathcal D}(\tau_e\mid s)\,
        p_{\mathcal D}(\tau_a\mid s)
    }
    \nonumber\\
    &=
    \log
    \frac{
        p_{\mathcal D}(\tau_a\mid s,\tau_e)
    }{
        p_{\mathcal D}(\tau_a\mid s)
    },
    \label{eq:app_optimal_interaction}
\end{align}
where the final equality follows from
\begin{equation}
    p_{\mathcal D}(\tau_e,\tau_a\mid s)
    =
    p_{\mathcal D}(\tau_e\mid s)\,
    p_{\mathcal D}(\tau_a\mid s,\tau_e).
\end{equation}

Now consider a logged interaction
$(s,\tau_e^\beta,\tau_a^\beta)$ and a candidate ego trajectory
$\hat{\tau}_e$. Evaluating \eqref{eq:app_optimal_interaction} at the
logged and candidate ego trajectories while keeping the same
surrounding-agent trajectory gives
\begin{align}
    &
    g_I^*(s,\tau_e^\beta,\tau_a^\beta)
    -
    g_I^*(s,\hat{\tau}_e,\tau_a^\beta)
    \nonumber\\
    &=
    \log
    \frac{
        p_{\mathcal D}(\tau_a^\beta\mid s,\tau_e^\beta)
    }{
        p_{\mathcal D}(\tau_a^\beta\mid s)
    }
    -
    \log
    \frac{
        p_{\mathcal D}(\tau_a^\beta\mid s,\hat{\tau}_e)
    }{
        p_{\mathcal D}(\tau_a^\beta\mid s)
    }
    \nonumber\\
    &=
    \log
    \frac{
        p_{\mathcal D}(\tau_a^\beta\mid s,\tau_e^\beta)
    }{
        p_{\mathcal D}(\tau_a^\beta\mid s,\hat{\tau}_e)
    }
    \nonumber\\
    &=
    \Delta_I.
\end{align}
The common marginal
$p_{\mathcal D}(\tau_a^\beta\mid s)$ therefore cancels, proving that the
difference of residual interaction scores exactly recovers IDS at the
population optimum.
\hfill$\square$

\paragraph{Why the candidate distribution must be shared.}
The use of the same candidate distribution in both contrastive objectives
is essential to the cancellation above. To see this, suppose instead that
the joint and ego classifiers used candidate distributions
$\nu_J(\tau_e\mid s)$ and $\nu_E(\tau_e\mid s)$, respectively. Their
population-optimal difference would become
\begin{equation}
    g_J^* - g_E^*
    =
    \log
    \frac{
        p_{\mathcal D}(\tau_a\mid s,\tau_e)
    }{
        p_{\mathcal D}(\tau_a\mid s)
    }
    +
    \log
    \frac{
        \nu_E(\tau_e\mid s)
    }{
        \nu_J(\tau_e\mid s)
    }.
\end{equation}
The second term introduces an additional proposal-dependent contribution.
Using the same policy-generated candidates for both objectives removes
this term and isolates the interaction-specific density ratio.

\subsection{KL Interpretation of Interaction Distribution Shift}
\label{app:ids_kl}

For clarity, define the pointwise interaction shift for an arbitrary
surrounding-agent trajectory $\tau_a$ as
\begin{equation}
    \Delta_I
    (s,\tau_e^\beta,\hat{\tau}_e,\tau_a)
    =
    \log
    \frac{
        p_{\mathcal D}(\tau_a\mid s,\tau_e^\beta)
    }{
        p_{\mathcal D}(\tau_a\mid s,\hat{\tau}_e)
    }.
    \label{eq:app_pointwise_ids}
\end{equation}

For fixed $s$, $\tau_e^\beta$, and $\hat{\tau}_e$, its expectation under
the conditional surrounding-agent distribution associated with the logged
ego trajectory is
\begin{align}
    &
    \mathbb E_{\tau_a\sim
        p_{\mathcal D}(\cdot\mid s,\tau_e^\beta)}
    \left[
        \Delta_I
        (s,\tau_e^\beta,\hat{\tau}_e,\tau_a)
    \right]
    \nonumber\\
    &=
    \mathbb E_{\tau_a\sim
        p_{\mathcal D}(\cdot\mid s,\tau_e^\beta)}
    \left[
        \log
        \frac{
            p_{\mathcal D}(\tau_a\mid s,\tau_e^\beta)
        }{
            p_{\mathcal D}(\tau_a\mid s,\hat{\tau}_e)
        }
    \right]
    \nonumber\\
    &=
    D_{\mathrm{KL}}
    \left(
        p_{\mathcal D}(\tau_a\mid s,\tau_e^\beta)
        \,\middle\|\,
        p_{\mathcal D}(\tau_a\mid s,\hat{\tau}_e)
    \right).
    \label{eq:app_ids_kl}
\end{align}

Hence, the pointwise IDS measures the change in support assigned to a
particular surrounding-agent realization, while its expectation measures
the divergence between the full conditional interaction distributions
associated with the logged and candidate ego trajectories.
\hfill$\square$

\newpage

\section{Implementation Details}
\label{app:implementation}

This section provides additional details on the data construction,
network architectures, and optimization procedure used for
\textsc{ICDP}.

\subsection{Data and Trajectory Representation}
\label{app:data}

We construct chunk-level transitions from the nuPlan trainval logs. Each
transition contains the current scene $s$, a logged ego trajectory
$\tau_e^\beta$, the corresponding surrounding-agent futures
$\tau_a^\beta$, the discounted return over the action horizon, and the
subsequent scene. Ego trajectories contain $32$ future steps sampled at
$10\,\mathrm{Hz}$, corresponding to a $3.2\,\mathrm{s}$ planning horizon.
The scene history contains $16$ frames spanning the preceding
$1.5\,\mathrm{s}$. Trajectories are represented in the current ego
rear-axle frame, while the subsequent scene is independently expressed
in the ego frame at the end of the action chunk.

The processed dataset contains $744{,}832$ transitions spanning $62$
recorded scenario types, including $12{,}108$ recovery transitions.
We reserve approximately $0.5\%$ of the data using a deterministic split,
resulting in $741{,}172$ training and $3{,}660$ validation transitions.
The resulting scenario distribution is naturally imbalanced and is shown
in Fig.~\ref{fig:scenario-distribution}.

The actor and critic scene representation contains agent histories,
vectorized map elements, preferred-route information, and static objects.
We retain up to $32$ agents and $5$ static objects for the actor--critic
scene encoder, together with at most $70$ lane and $25$ route polylines.
Each polyline is represented by $20$ points within a $100\,\mathrm{m}$
map-query radius. Ego actions are represented as
$(x,y,\cos\psi,\sin\psi)$ over the prediction horizon. For the
interaction-support estimator, the final \textsc{ICDP} configuration uses
up to $48$ surrounding-agent trajectories.

\subsection{Reward Function}
\label{app:reward}

Each ego trajectory is evaluated over an action chunk of $H=32$ steps
at $\Delta t=0.1$\,s. The reward combines normalized route progress with
multiplicative factors for route compliance, lane alignment, speed, time to
collision (TTC), and comfort. Additional penalties are applied for
near-critical drivable-area and TTC violations.

Let $\mathcal{F}$ denote the occurrence of either an ego collision or a
drivable-area violation larger than $0.5$\,m at any point in the chunk.
The chunk return is
\begin{equation}
    R =
    \begin{cases}
        -5, & \text{if } \mathcal{F},\\[2pt]
        \displaystyle
        \sum_{h=1}^{H}\gamma^{h-1} r_h,
        & \text{otherwise},
    \end{cases}
    \qquad \gamma=0.99,
    \label{eq:reward-chunk-return}
\end{equation}
where an invalidating event also disables bootstrapping. For valid
trajectories, the per-step reward takes the form
\begin{equation}
    r_h
    =
    p_h
    \prod_{k\in\mathcal{K}} f_h^k
    -
    c_h^{\mathrm{area}}
    -
    c_h^{\mathrm{ttc}},
    \qquad
    \mathcal{K}
    =
    \{
    \mathrm{area},
    \mathrm{center},
    \mathrm{direction},
    \mathrm{speed},
    \mathrm{ttc},
    \mathrm{comfort}
    \},
    \label{eq:reward-step}
\end{equation}
where $p_h$ denotes speed-normalized progress along the preferred route.
Backward progress is clipped to zero, while progress above the local
speed-limit distance is capped. The multiplicative factors remain in
$[0.1,1]$ (except comfort, which is lower bounded by $0.5$), so poor
driving quality reduces the reward without reversing the progress signal.

Table~\ref{tab:reward-parameters} summarizes the principal reward
parameters. TTC and collision checks are computed against all tracked
dynamic and static objects available in the logged scene. The ego- and
interaction-support penalties used by \textsc{ICDP} are applied separately
during policy optimization and are not part of the reward.

\begin{table}[t]
\centering
\caption{Principal reward parameters used for training.}
\label{tab:reward-parameters}

\small
\setlength{\tabcolsep}{7pt}
\renewcommand{\arraystretch}{1.08}

\begin{tabular}{lc}
\toprule
\textbf{Parameter} & \textbf{Value} \\
\midrule
Action horizon $H$ & $32$ steps ($3.2$\,s) \\
Sampling interval $\Delta t$ & $0.1$\,s \\
Discount factor $\gamma$ & $0.99$ \\
Collision / severe off-road return & $-5$ \\
Soft / hard drivable-area threshold & $0.3$ / $0.5$\,m \\
TTC additive-penalty threshold & $0.95$\,s \\
TTC factor saturation range & $0.5$--$3.0$\,s \\
Speed-limit tolerance & $0.5$\,m/s \\
Default speed limit & $13.9$\,m/s \\
Lane-direction tolerance & $20^\circ$ \\
Comfort-factor range & $[0.5,1]$ \\
\bottomrule
\end{tabular}
\end{table}

\subsection{Network Architecture}
\label{app:networks}

\paragraph{Scene encoding.}
The actor and critic use the same scene-encoder architecture with separate
parameters. Agent histories and lane polylines are processed using
modality-specific MLP-Mixer encoders, while static objects are embedded
with an MLP. Position and semantic-type embeddings are added before
multi-modal fusion through two Transformer layers with hidden dimension
$192$ and eight attention heads. The critic encoder is initialized from
the actor encoder and subsequently optimized independently.

\paragraph{Diffusion actor.}
The actor predicts the complete clean ego trajectory using an
$x_0$ diffusion parameterization. The current ego state and noisy future
trajectory are embedded into a trajectory-query token, which cross-attends
to the encoded scene through two diffusion-Transformer layers. The decoder
is additionally conditioned on the preferred route and diffusion time.
Only the ego trajectory is generated; surrounding-agent trajectories are
used as observed context rather than prediction targets.

\paragraph{Critic ensemble.}
The critic evaluates clean candidate trajectories. The $32$ trajectory
steps are first embedded as time-indexed query tokens and processed by a
two-layer trajectory trunk using temporal self-attention and cross-attention
to the scene representation. The resulting trajectory representation is
combined with pooled scene features and the normalized action sequence.
Five independent MLP value heads then predict the chunk-level action value.
The pessimistic estimate used for policy improvement is obtained from the
ensemble mean and standard deviation as defined in~\eqref{eq:q_lcb}.

\paragraph{Interaction-support networks.}
The support estimator reuses pooled features from the delayed critic,
with gradients stopped before the support networks. The ego trajectory
and scene state are separately projected to $128$-dimensional features.
For the joint classifier, surrounding-agent trajectories, their motion
relative to the candidate ego trajectory, and trajectory-validity masks
are additionally encoded and aggregated using masked mean and max pooling.
The ego and joint classifiers use three-layer MLP heads with hidden width
$256$. Their outputs are the logits $g_E$ and $g_J$; the residual
interaction score $g_I=g_J-g_E$ is computed directly and is not
parameterized by a third network.

\subsection{Learning and Optimization}
\label{app:optimization}

\paragraph{Reward and chunk-level critic learning.}
The chunk reward combines progress along the preferred route with terms
capturing collision avoidance, time to collision, drivable-area and route
compliance, lane positioning, travel direction, speed, and comfort.
Collisions and severe departures from the drivable area invalidate the
action chunk, assign a terminal penalty, and disable bootstrapping.

We use a per-step discount of $\gamma=0.99$. For valid transitions, four
candidate trajectories are sampled from the target policy at the subsequent
state, and their target-critic lower-confidence-bound estimates are averaged:
\begin{equation}
    y
    =
    R^{(H)}
    +
    \gamma^H m
    \frac{1}{K}
    \sum_{k=1}^{K}
    Q_{\mathrm{LCB}}^{\mathrm{targ}}
    (s',\tau'_{e,k}),
    \qquad K=4,
    \label{eq:appendix_bellman}
\end{equation}
where $m$ denotes the bootstrap-validity mask. All critic heads regress
toward the same detached target.

\paragraph{Diffusion-policy optimization.}
The actor uses a variance-preserving diffusion process with a linear
noise schedule $\beta(t)=0.1+19.9t$ and
$t\sim\mathcal U(10^{-3},1)$. During policy optimization, four candidate
trajectories are decoded from each noisy action using five reverse-diffusion
steps. Candidate value and support estimates are evaluated on the
corresponding clean trajectories and used to construct the noisy-space
policy gradient. The diffusion denoising objective remains active throughout
training and acts as a behavioral anchor. Reward-driven guidance and
interaction-support regularization are enabled after a $5{,}000$-update
warm-up.

\paragraph{Contrastive support learning.}
The ego and joint classifiers use the same policy-generated candidate
trajectories as negative examples. Logged ego trajectories define the
positive class, while logged surrounding-agent futures remain fixed when
constructing joint examples. Both classifiers are trained with balanced
binary cross-entropy, and delayed copies are used when scoring actor
candidates. The final policy configuration uses
$\lambda_E=0.3$ and $\lambda_I=1$, as analyzed in
Sec.~\ref{sec:interaction_analysis}.

\paragraph{Optimization.}
Actor, critic, and support networks use separate AdamW optimizers with
learning rate $3\times10^{-4}$, zero weight decay, and gradient-norm
clipping at $1.0$. Training uses four distributed workers with $256$
transitions per worker, giving a global batch size of $1{,}024$. Target
networks are updated by exponential moving average with coefficient
$0.005$. Unless otherwise stated, training is run for up to $250{,}000$
updates.

\paragraph{Recovery and actor augmentation.}
To expose the policy to recoverable deviations from the demonstrations,
we additionally construct recovery trajectories by perturbing eligible ego
states and reconnecting them to the logged future after $2.0\,\mathrm{s}$.
During actor training, eligible states are further perturbed within bounded
lateral and heading offsets and the initial portion of the denoising target
is refitted accordingly. Perturbed examples contribute to the diffusion
anchor but are excluded from value guidance and contrastive support updates,
ensuring that the interaction objectives remain defined with respect to
consistent logged scenes and surrounding-agent futures.

\begin{table}[t]
\centering
\caption{
Principal implementation settings used for \textsc{ICDP}.
}
\label{tab:implementation-parameters}

\small
\setlength{\tabcolsep}{7pt}
\renewcommand{\arraystretch}{1.08}

\begin{tabular}{lc}
\toprule
\textbf{Parameter} & \textbf{Value} \\
\midrule
Planning horizon & $32$ steps ($3.2\,\mathrm{s}$) \\
History horizon & $16$ frames ($1.5\,\mathrm{s}$) \\
Actor/critic agent capacity & $32$ \\
Interaction-support agent context & $48$ \\
Lane / route capacity & $70$ / $25$ \\
Map-query radius & $100\,\mathrm{m}$ \\
Scene hidden dimension & $192$ \\
Scene / actor Transformer layers & $2$ / $2$ \\
Critic ensemble size & $5$ \\
Support feature dimension & $128$ \\
Per-step discount $\gamma$ & $0.99$ \\
Pessimism weight $\kappa_Q$ & 1 \\
Bellman-target samples & $4$ \\
NSPG candidate samples & $4$ \\
Reverse diffusion steps & $5$ \\
Optimizer & AdamW \\
Learning rate & $3\times10^{-4}$ \\
Global batch size & $1{,}024$ \\
EMA coefficient & $0.005$ \\
$\lambda_E$ / $\lambda_I$ / $\lambda_{BC}$ & $0.3$ / $1.0$ / $1.0$ \\
Guidance warm-up & $5{,}000$ updates \\
Training budget & $250{,}000$ updates \\
\bottomrule
\end{tabular}
\end{table}

\subsection{Qualitative Samples and Dataset Composition}
\label{app:qualitative}

\begin{figure*}[p]
    \centering

    \begin{minipage}[t]{0.48\textwidth}
        \centering
        \includegraphics[
            width=0.88\linewidth,
            keepaspectratio
        ]{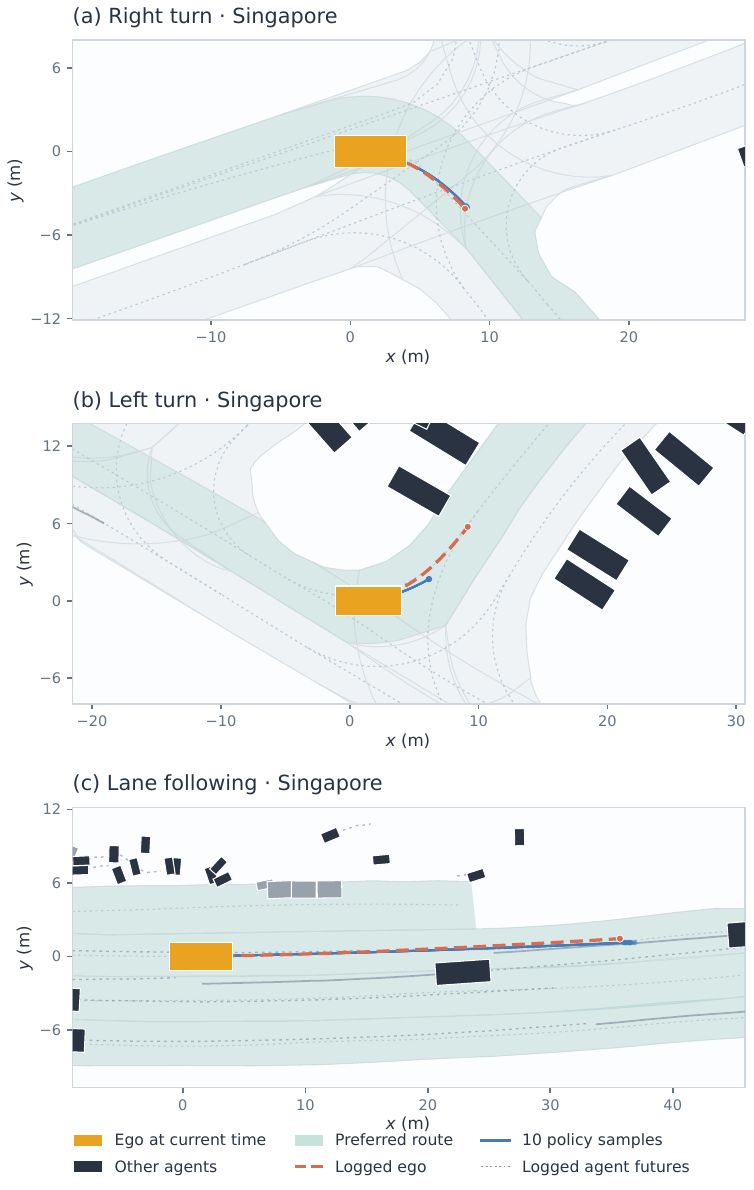}
        \small (a) Singapore
    \end{minipage}
    \hfill
    \begin{minipage}[t]{0.48\textwidth}
        \centering
        \includegraphics[
            width=0.88\linewidth,
            keepaspectratio
        ]{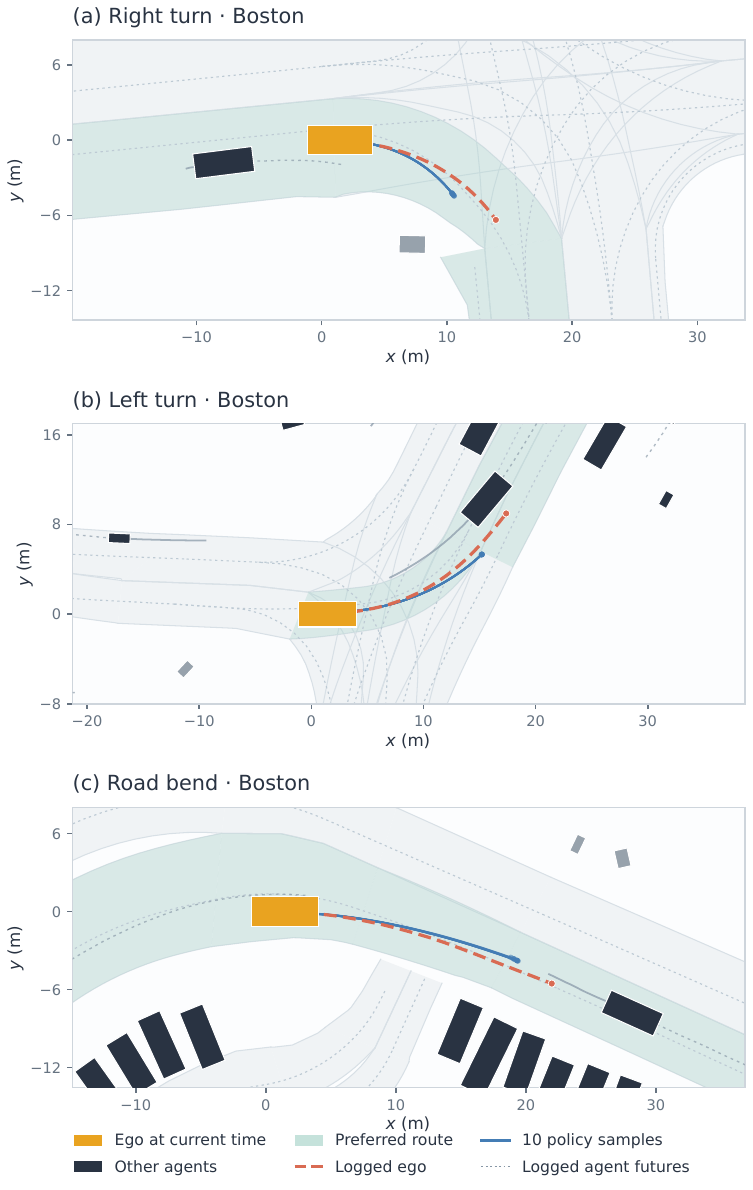}
        \small (b) Boston
    \end{minipage}

    \vspace{6pt}

    \begin{minipage}[t]{0.48\textwidth}
        \centering
        \includegraphics[
            width=0.88\linewidth,
            keepaspectratio
        ]{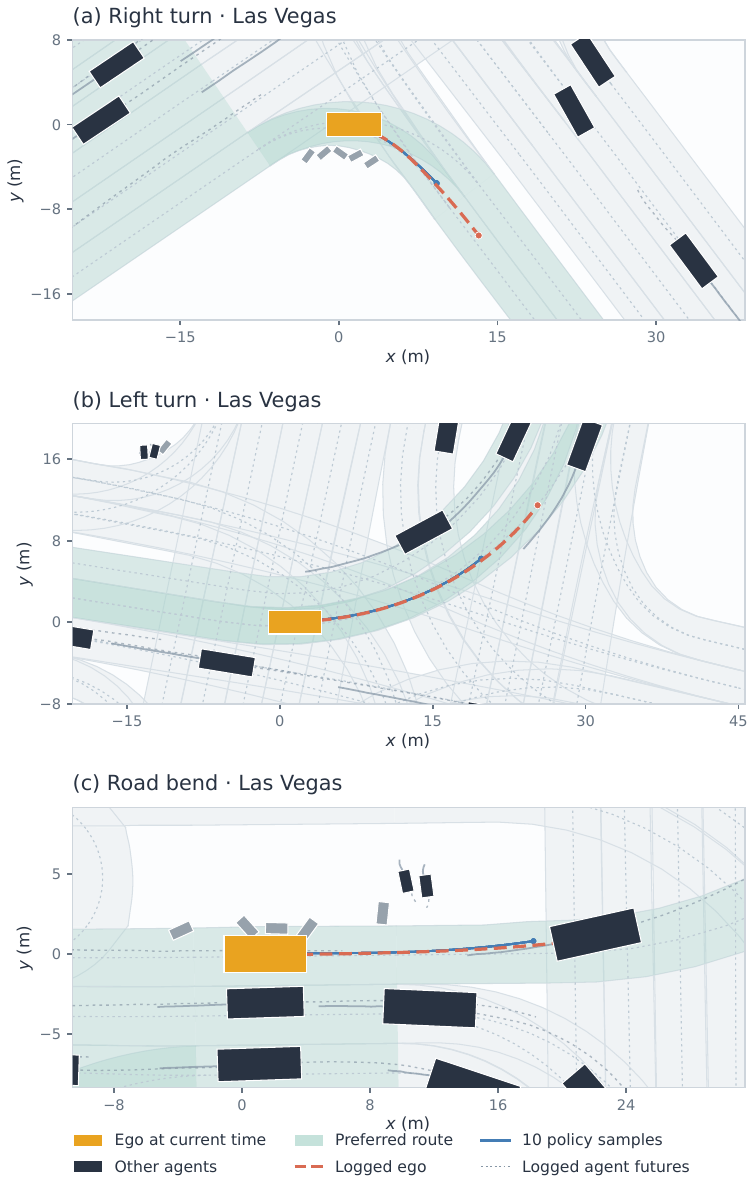}
        \small (c) Las Vegas
    \end{minipage}
    \hfill
    \begin{minipage}[t]{0.48\textwidth}
        \centering
        \includegraphics[
            width=0.88\linewidth,
            keepaspectratio
        ]{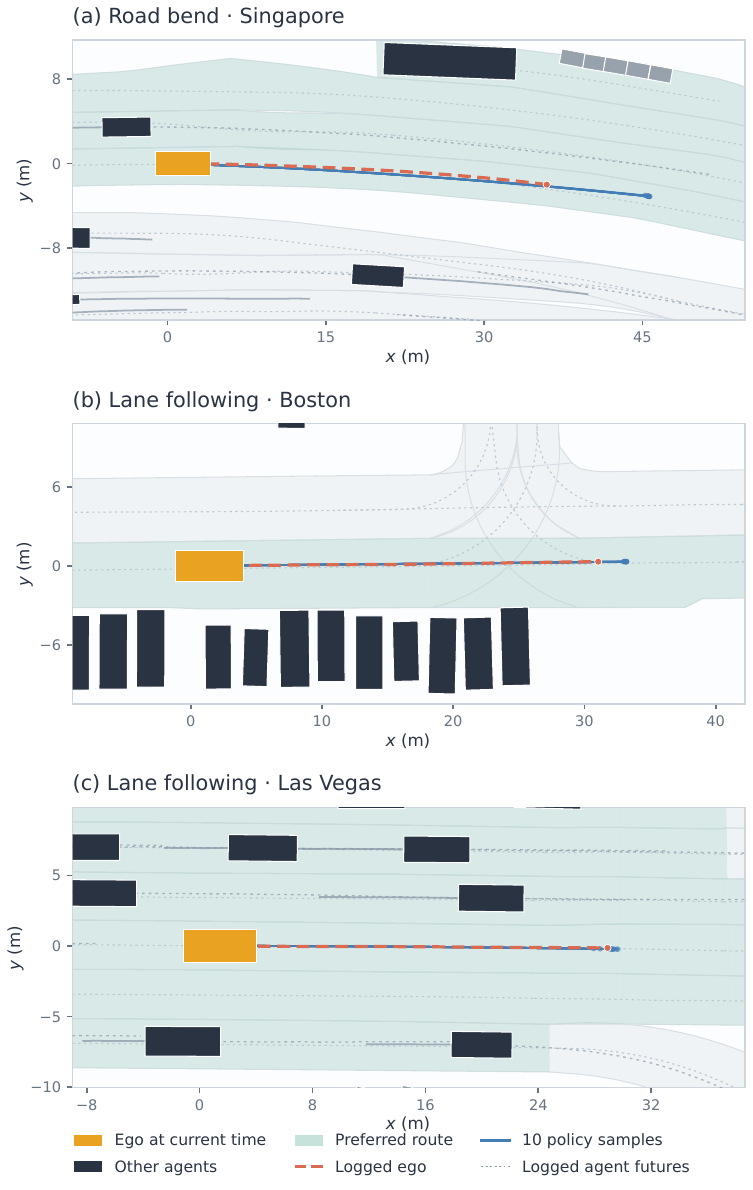}
        \small (d) Additional cross-city scenes
    \end{minipage}

    \caption{
    Qualitative trajectory samples from the learned diffusion policy across
    diverse nuPlan scenes. Each panel shows ten independently generated ego
    trajectories together with the logged ego trajectory, preferred route,
    and surrounding traffic. Samples are shown directly without critic-based
    selection or post-processing.
    }
    \label{fig:qual_samples}
\end{figure*}

\paragraph{Policy samples.}
Figure~\ref{fig:qual_samples} visualizes trajectory samples from the learned
diffusion policy across diverse road geometries and traffic configurations.
The examples span Singapore, Boston, and Las Vegas, covering turning,
lane-following, and curved-road scenarios. For each scene, ten trajectories
are sampled independently from the EMA policy and shown without critic-based
ranking, rejection, smoothing, or post-processing. The logged ego trajectory
and surrounding-agent futures are included for reference. These examples
illustrate the diversity and geometric consistency of the learned trajectory
distribution rather than closed-loop performance.

\begin{figure}[t]
    \centering

    \begin{minipage}[t]{0.49\linewidth}
        \centering
        \includegraphics[width=\linewidth]
        {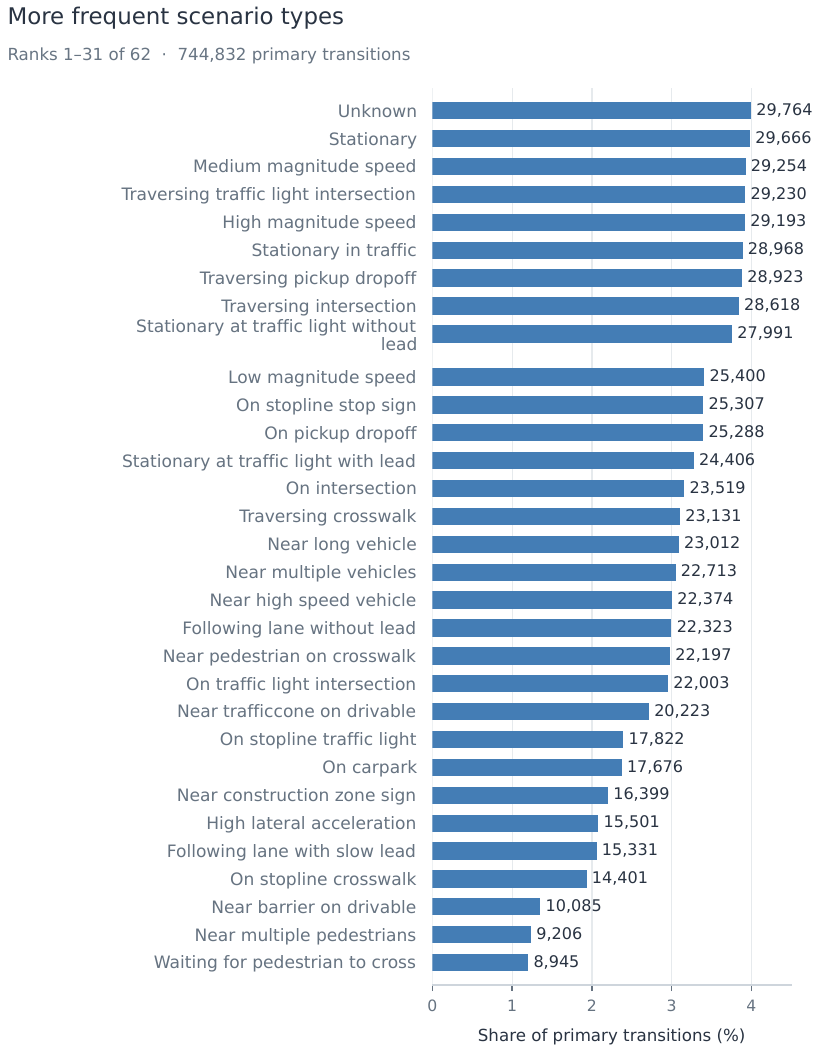}
        \small (a) More frequent scenario types
    \end{minipage}
    \hfill
    \begin{minipage}[t]{0.49\linewidth}
        \centering
        \includegraphics[width=\linewidth]
        {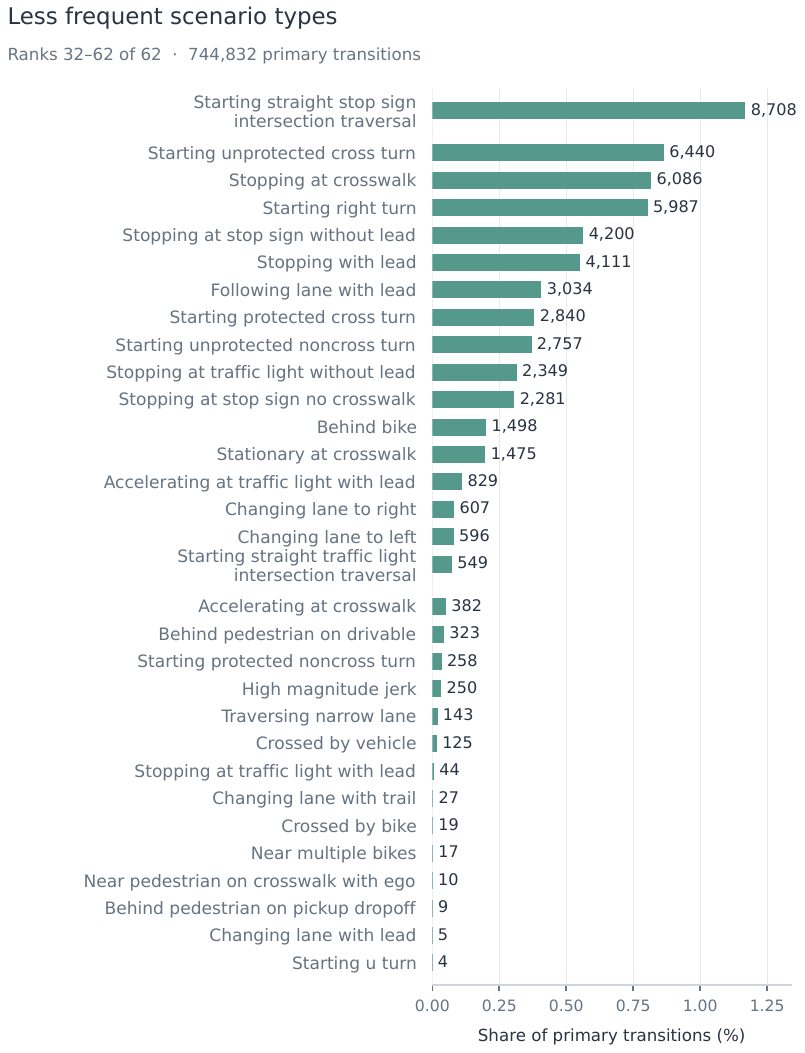}
        \small (b) Less frequent scenario types
    \end{minipage}

    \caption{
    Distribution of scenario types in the processed offline dataset.
    The $62$ recorded categories are divided into more- and less-frequent
    groups to make the long-tailed distribution legible.
    }
    \label{fig:scenario-distribution}
\end{figure}

\paragraph{Dataset composition.}
The offline dataset comprises $62$ nuPlan scenario categories, representing a diverse range of driving situations with a long-tailed frequency distribution. Figure~\ref{fig:scenario-distribution} illustrates the empirical distribution across all categories, with the more- and less-frequent halves presented separately to improve readability.

\begin{table}[t]
\centering
\caption{
Closed-loop performance on InterPlan and selected interaction-critical
scenario categories. The best result in each column is highlighted in blue.
}
\label{tab:interplan}

\small
\setlength{\tabcolsep}{10pt}
\renewcommand{\arraystretch}{1.12}

\begin{tabular}{lcccc}
\toprule
\textbf{Planner}
& \textbf{Overall}
& \textbf{Nudge Around}
& \textbf{High Traffic}
& \textbf{Jaywalk} \\
\midrule

PlanTF
& 47.70
& 49.40
& 58.85
& 33.94 \\

PLUTO w/o refine.
& 58.47
& 71.56
& 67.25
& 25.48 \\

Diffusion Planner
& 52.90
& 60.48
& 49.71
& 26.20 \\

Flow Planner
& 61.82
& \cellcolor{bestblue}72.96
& 67.21
& 43.57 \\

\midrule

\textsc{ICDP}
& \cellcolor{bestblue}62.7
& 51.8
& \cellcolor{bestblue}70.2
& \cellcolor{bestblue}51.1 \\

\bottomrule
\end{tabular}
\end{table}

\section{Additional InterPlan Evaluation}
\label{app:interplan}

We additionally evaluate \textsc{ICDP} on InterPlan, which emphasizes
interactive and challenging closed-loop driving scenarios. We compare
against the same representative planning methods where reported results are
available. Table~\ref{tab:interplan} reports the overall score together with
three interaction-focused scenario categories.


\end{document}